\documentclass[journal]{IEEEtran}
\usepackage[numbers]{natbib}

\usepackage{verbatim}
\usepackage{booktabs} 
\usepackage{amsmath}
\usepackage{multicol, blindtext}
\usepackage{algorithm}
\usepackage[noend]{algpseudocode}
\usepackage{courier}
\usepackage{amsmath}
\usepackage{graphicx}
\makeatletter
\def\BState{\State\hskip-\ALG@thistlm}
\makeatother
\usepackage{caption,subcaption}
\usepackage{textcomp}
\usepackage{multirow}
\usepackage[latin1]{inputenc}
\usepackage{tikz}
\usetikzlibrary{shapes,arrows}
\tikzstyle{decision} = [diamond, draw, fill=white!20, 
    text width=4.5em, text badly centered, node distance=3cm, inner sep=0pt]
\tikzstyle{block_init} = [rectangle, draw, fill=white!20, 
    text width=9em, text centered, rounded corners, minimum height=2em]
\tikzstyle{block} = [rectangle, draw, fill=white!20, 
    text width=5em, text centered, rounded corners, minimum height=2em]
\tikzstyle{cblock} = [rectangle, draw, fill=gray!20, 
    text width=5em, text centered, rounded corners, minimum height=2em]
\tikzstyle{line} = [draw, -latex']
\tikzstyle{cloud} = [draw, ellipse,fill=white!20, node distance=3cm,
    minimum height=2em]
\usepackage{mathtools}

\newcommand{\boldx}{\mbox{${\mathbf x}$}}

\ifCLASSINFOpdf
\else
\fi
\begin{document}
%
\title{Enhanced \textit{Innovized} Repair Operator for Evolutionary Multi- and Many-objective Optimization}
%
%
%

\author{Sukrit~Mittal,
        Dhish~Kumar~Saxena, Kalyanmoy~Deb,
        and~Erik~Goodman
        \\ {\bf COIN Report Number 2020020}
\thanks{S. Mittal and D.K. Saxena are with the Department
of Mechanical and Industrial Engineering, Indian Institute of Technology, Roorkee,
Uttarakhand, 247667 India, e-mail: \{smittal1,dhish.saxena\}@me.iitr.ac.in.}
\thanks{K. Deb and E. Goodman are with the Beacon NSF Center and the Department of Electrical and Computer Engineering, Michigan State University, East Lansing,
MI, 48824 USA, e-mail: \{kdeb,goodman\}@egr.msu.edu.}
}

\newcommand{\eg}[1]{{\color{green}{#1}}}

\maketitle

\begin{abstract}
``Innovization'' is a task of learning common relationships among some or all of the Pareto-optimal (PO) solutions in multi- and many-objective optimization problems. Recent studies have shown that a chronological sequence of non-dominated solutions obtained in consecutive iterations during an optimization run also possess salient patterns that can be used to learn problem features to help create new and improved solutions. In this paper, we propose a machine-learning- (ML-) assisted modeling approach that learns the modifications in design variables needed to advance population members towards the Pareto-optimal set. We then propose to use the resulting ML model as an additional \textit{innovized} repair (IR2) operator to be applied on offspring solutions created by the usual genetic operators, as a novel mean of improving their convergence properties. In this paper, the well-known random forest (RF) method is used as the ML model and is integrated with various evolutionary multi- and many-objective optimization algorithms, including NSGA-II, NSGA-III, and MOEA/D. On several test problems ranging from two to five objectives, we demonstrate improvement in convergence behavior using the proposed IR2-RF operator. Since the operator does not demand any additional solution evaluations, instead using the history of gradual and progressive improvements in solutions over generations, the proposed ML-based optimization opens up a new direction of optimization algorithm development with advances in AI and ML approaches. 
\end{abstract}

\begin{IEEEkeywords}
Multiobjective Optimization, Learning-based Optimization, Random Forest, Machine Learning, \textit{Innovization}, Online \textit{Innovization}
\end{IEEEkeywords}

\IEEEpeerreviewmaketitle

\section{Introduction}

\IEEEPARstart{O}{ptimization} is an iterative process of arriving one or more optimal or near-optimal solutions for single- and multi-objective optimization problems using a point-based or a population-based approach. For multi- (having two or three objectives) and many-objective (having more than three objectives) optimization, the target is usually a set of trade-off Pareto-optimal (PO) solutions. Population-based methods become ideal for identifying a set of well-distributed and well-converged solutions in such problems. Since evolutionary algorithms use a population of solutions in each iteration, their extensions -- evolutionary multi-objective optimization (EMO) or evolutionary many-objective optimization (EMaO) algorithms -- are extensively used for solving them \cite{deb-book-01,carlos-book}. 

EMO/EMaO algorithms, including NSGA-II \cite{deb2002nsga1}, NSGA-III \cite{nsga3-1,nsga3-2}, MOEA/D \cite{moead}, create new solutions (called offspring solutions) using bio-inspired genetic operators, like recombination and mutation operators. However, there are other algorithms, such as estimation of distribution algorithms (EDAs), that adopt a more involved approach and create new solutions by performing a sampling procedure from either a custom-built probabilistic model or other available probability models, such as Bayesian Networks, decision-trees, etc. EDAs, like BOA \cite{boa} and MONEDA \cite{MONEDA}, have shown that there is a potentially distinctive advantage in using the inter-variable relationships to generate new offspring solutions. This linkage-preserving advantage may be more important in problems with highly-linked variables. However, there are other problems where independent mating of variables through operators like SBX crossover causes a more efficient search \cite{sukrit-coin-ann}.

Some recent studies in the EMO/EMaO domain have focused on an \textit{online innovization} approach \cite{abhinav-cec17,deb-datta-ejor,rocket-cec20} which attempts to extract decision variable patterns or relationships that exist among the current non-dominated solutions and then apply this information as additional operators (apart from the crossover and mutation operators) in subsequent generations. These techniques explore the advantage of independent variable mating and exploit the learned inter-variable relationships at the same time. First, these studies have primarily been restricted to specific pre-fixed structures of relationships, such as power laws  \cite{eopaper,dudas2013integration}. Some limited studies of a more generic nature, using a genetic programming approach, have also been performed \cite{bandaru-gp}, which provide a promising direction for exploration and exploitation of these inter-variable relationships. 

However, the scope of online innovization concept can also also be extended to utilize the history of non-dominated solutions from earlier generations of an EMO/EMaO run to construct a {\em mapping model\/} which can improve an arbitrary variable-vector to map it closer to the current non-dominated set. A recent study \cite{sukrit-ann-repair-gecco,sukrit-coin-ann} has used artificial neural networks (ANNs) to learn a generic variable mapping between dominated and currently non-dominated target solutions and used the trained ANN model to repair the offspring solutions, thus providing offspring solutions to progress in the direction of the current non-dominated set. This learning process is discussed in Section~\ref{sec:ANN-repair}.

In this paper, we propose a random forest (RF) assisted machine learning (ML) method to capture the directional improvements in the optimization problem's decision space in a progressive manner along with a few other improvements. Linking past dominated solutions with a specific target non-dominated solution, an RF is trained to develop a model, which is then used within the EMO/EMaO run to repair the generated offspring solutions into the \textit{learned} directions. The proposed ML method is designed to focus on maintaining a diverse solution set at all times during the EMO/EMaO run, so that in addition to the convergence properties, a well-diverse set of solutions are also emphasized. The earlier naive approach \cite{sukrit-ann-repair-gecco} executed with ANN was reported to have difficulties in making useful repairs in certain problems. The proposed \textit{innovized} repair operator with RF method (IR-RF) is reported to perform better on various problems covering different test suites, including ZDT, DTLZ, and WFG, with problems ranging from two to five objectives.  
Besides using RF over ANN, the basic IR operator is also modified with a number of other extensions which has produced an improved IR (IR2) operator. Extensive comparative evaluation of the proposed IR2-RF is made to reveal the benefits of the innovized repair operator concept. 


In the remainder of this paper, we provide a brief overview of the existing learning-assisted EMO/EMaO algorithms including state-of-the-art methods for \textit{online innovization}, and the past ANN-assisted repair operators in Sections~\ref{sec:innovization} and \ref{sec:ANN-repair}, respectively. The proposed IR-RF operator is outlined in Section~\ref{sec:IR-RF}, followed by some comparative results. Section~\ref{sec:proposed} discusses modifications to IR operator by augmenting it with a number of specific features that allow the proposed IR2 operator to yield performance improvements. The differences between IR and IR2 operators are discussed in Section~\ref{sec:differences}. Section~\ref{sec:results} presents the results with new IR2 operator on a wide range of test problems, followed by a discussion of results in Section~\ref{sec:discussion}. The paper is concluded in Section~\ref{sec:conclusions}.

\section{Existing State-of-the-Art} \label{sec:background}

In EMO/EMaO algorithms, the genetic operators---i.e., crossover and mutation---are not effective in reaching the optimal solution quickly for certain problems \cite{survey_prob, eda_svm}. Towards this goal, some approaches have been coupled with EMO/EMaO algorithms, such as local-search methods and surrogate-modeling techniques, to reduce the overall number of function evaluations. A recent study \cite{sukrit-coin-ann} discusses how an ANN-assisted repair operator that builds a  foundation for the proposed repair operator is inherently different from local-search or surrogate methods. Further, there are Hybrid Genetic Algorithms (GAs) that learn from the evolution in past iterations and use that information to guide the evolution of new solutions, further discussed in the next paragraph. Apart from these, sections \ref{sec:innovization} and \ref{sec:ANN-repair} discuss the idea of \textit{innovized repair} and the ANN-assisted repair operator.

Hybrid Approaches with repair heuristics can be broadly classified into Lamarckian and Baldwinian approaches \cite{repair-survey}. If the genotype or the decision variable vector is changed using a local repair heuristic, it is known as a Lamarckian algorithm. On the other hand, Baldwinian-learning-based heuristics modify the fitness landscape rather than the genotype. \cite{memetic-de} suggests that adaptive Baldwinian learning can do well in guiding the course of evolution in an EA, altering the shape of the search space and thus providing good evolutionary paths for the individuals. It may be noted that Baldwinian Learning is not a learning mechanism or method; rather, it designates the manner of learning. An example of this is the Multiobjective Immune Algorithm with Baldwinian Learning (MIAB) \cite{MIAB}. In MIAB, the learning operator at generation $t$ captures the individual evolving directions (in decision-space) from $t-1$ to $t$ and uses it to decide the individual direction of evolution 
from $t$ to $t+1$. It may be noted that this learning-based evolution operator is used as an alternative to genetic operators---i.e., crossover and mutation---but our proposed approach repairs/evolves the offspring which have already been produced using genetic operators.  


\subsection{Innovization} \label{sec:innovization}
\textit{Innovization} is a knowledge-discovery task first proposed by Deb \cite{deb-design} and was later exemplified on several engineering design problems \cite{deb-innovization}. Even though \textit{innovization} was proposed initially as a post-optimization process, a few studies \cite{abhinav-cec17,rocket-cec20} later claimed that knowing these relationships through some intermediate-generation solutions during an optimization run and using this knowledge can enhance the convergence of the optimization algorithm. These relationships can be used to \textit{repair} the solutions generated by EMO/EMaO operators, thus forcing these new solutions to possess some properties obtained from those good solutions, eventually accelerating the convergence. There are many issues and variables regarding learning at intermediate generations, such as the process/mechanism of learning, type/format of information to be extracted, and how this extracted information can be used to improve the existing solutions. 

\subsection{ANN-assisted Innovized Repair (IR-ANN) Operator} \label{sec:ANN-repair}
The study in \cite{sukrit-ann-repair-gecco} made the first attempt to capture the direction of evolution in a multi-dimensional search space using an ANN and applied it back as an offspring repair operator to enhance the speed of convergence. A more comprehensive study on the same IR operator was done in \cite{sukrit-coin-ann}. This IR operator has three processing steps, i.e., (i) Generation of training-dataset (in normal EMO/EMaO process), (ii) ANN-Training, and (iii) Repairing offspring solutions. Any intermediate generation $t$ of the EMO/EMaO with the ANN-assisted repair operator is described in the steps below.

\begin{description}
    \item[Step 1:] Update the Archive $A_t$ with population members from $t_{past}$ previous generations. 
    \item[Step 2: ] Check the population threshold, i.e., whether or not 50\% population from the current population $P_t$ is non-dominated. This is a deciding factor whether or not the learning and repair operation will happen at generation $t$.
    \item[Step 3: ] Generate the training-dataset $D_t$ from $A_t$. Towards this, every member in $A_t$ is associated to the nearest reference vector (say $Z_i$), and then is paired with the non-dominated solution (in current population $P_t$) nearest to that particular reference vector $Z_i$.
    \item[Step 4: ] Train an ANN using $D_t$.
    \item[Step 5: ] Generate offspring population $Q_t$ using the crossover and mutation operators on $P_t$. 
    \item[Step 6: ] Repair a randomly selected set of 50\% of the total offspring solutions $Q_t$ using the trained ANN model.
    \item[Step 7: ] Evaluate the offspring solutions.
    \item[Step 8: ] Select solutions from $P_t \cup Q_t$ to yield $P_{t+1}$.
\end{description}

The aforementioned process is guided by two user-defined parameters. The parameter $t_{past}$ governs how many previous generations are considered in the archive $A_t$ and the parameter $t_{freq}$ governs the frequency of repairs, i.e., the repair takes place only after $t_{freq}$ generations. The study \cite{sukrit-coin-ann} also discussed how the ANN-assisted repair operator differs from surrogate-modeling techniques or other local search operators for EMO/EMaO algorithms. 

\section{Innovized Repair (IR) Operator with RF} \label{sec:IR-RF}

In this section, we analyze some results for how the performance changes when the associated learning method is changed from ANN to random forest (RF) method, referred to as IR-ANN and IR-RF, respectively. Since the learning is in the decision space, each training input and output is a vector of length $n_{var}$ (number of variables). The RF has three critical parameters to be defined, i.e., the number of trees ($n_{trees}$), number of variables/features considered while splitting a node ($n_{features}$) and the splitting criterion. Considering that the size of the input archive is $N_{A}$, the following settings have been used for the RF: 
\begin{equation}
    n_{trees} = N_{A}, \quad n_{features} = n_{var}.
\end{equation}
In addition to this, the splitting criterion used is \textit{Mean Squared Error} (MSE), for a fair comparison with the ANN that uses MSE as the loss function. The rest of the implementation settings are taken as default\footnote{The Random Forest Regressor used in this study has been taken from this Scikit-learn implementation (for python language). \\ https://scikit-learn.org/stable/modules/generated/sklearn.ensemble.\\RandomForestRegressor.html}. Some comparison results are shown in Table~\ref{tab:ann-rf} when NSGA-II is used to solve ZDT problems with the IR operator (proposed earlier in \cite{sukrit-coin-ann} and discussed in Section~\ref{sec:ANN-repair}) but using ANN and RF as learning methods. These are median hypervolume values at the end of 200 generations. The better results are marked in bold. The parameters used for NSGA-II and the IR operator are discussed later in Section~\ref{sec:parameter-settings}.
\begin{table}[hbt] \label{tab:ann-rf}
\centering
\caption{Median Hypervolume on ZDT problems for NSGA-II with IR operator using ANN and RF as learning methods.}
\begin{tabular}{|c|c|c|c|} \hline
Problems & NSGA-II-IR-ANN & NSGA-II-IR-RF & $p$-value    \\ \hline
ZDT1     & 0.679193       & \textbf{0.680198}      & 8.98E-11 \\
ZDT2     & 0.345448       & \textbf{0.346683}      & 1.33E-11 \\
ZDT3     & 0.535027       & \textbf{0.535561}      & 1.62E-11 \\
ZDT4     & 0.680704       & \textbf{0.681041}      & 4.93E-04 \\
ZDT6     & \textbf{0.333949}       & 0.304714      & 1.40E-05 \\ \hline
\end{tabular}
\end{table}
This limited study indicates that there is a benefit of using RF over ANN in these kinds of problems. It may be noted that for ZDT6, even though RF performed worse than ANN, both NSGA-II-IR-ANN and NSGA-II-IR-RF did better than NSGA-II (HV = 0.296750). In addition to changing the learning method from ANN to RF, this paper proposes some modifications in the core of IR operator design like the training dataset generation step and the repair step, which are further discussed in next section.

\section{Modified Innovized Repair (IR2) Operator} \label{sec:proposed}
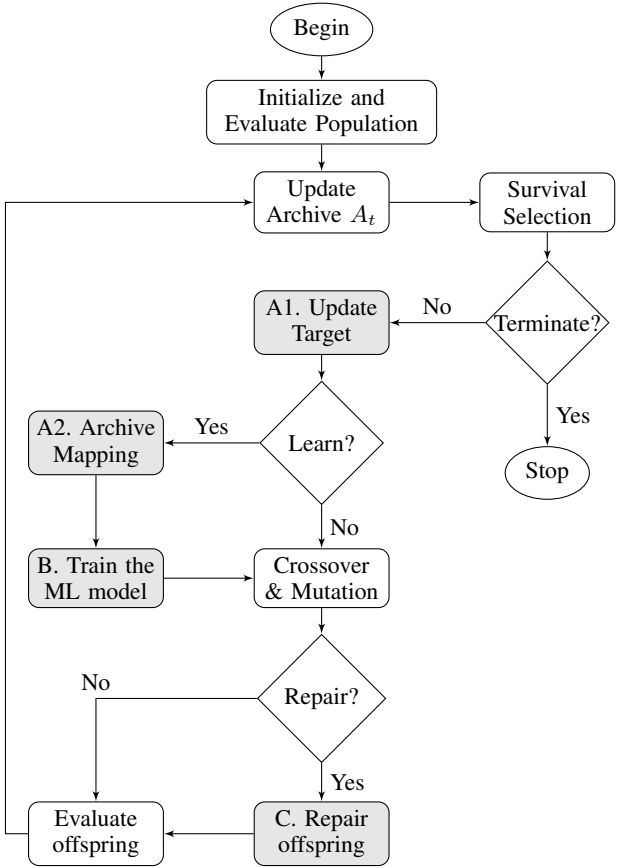
\begin{figure}[hbt]
\begin{tikzpicture}[node distance = 2cm, auto]
    \small
    \centering
    \node [cloud] (begin) {Begin};
    \node [block_init, below of=begin, yshift=0.9cm] (initialize) {Initialize and Evaluate Population};
    \node [block, below of=initialize, yshift=0.8cm] (uparchive) {Update Archive $A_t$};
    \node [block, right of=uparchive, xshift=1cm] (survival) {Survival Selection};
    \node[decision, below of=survival, yshift=1.4cm] (terminate) {Terminate?};
    \node [cloud, below of=terminate, yshift=1cm] (stop) {Stop};
    \node [cblock, left of=terminate, xshift=-1cm] (uptarget) {A1. Update Target};
    \node [decision, below of=uptarget, yshift=1.4cm] (learn) {Learn?};
    \node [cblock, left of=learn, xshift=-1cm] (mapping) {A2. Archive Mapping};
    \node [cblock, below of=mapping, yshift=0.2cm] (train) {B. Train the ML model};
    \node [block, below of=learn, yshift=0.2cm] (mating) {Crossover \& Mutation};
    \node [decision, below of=mating, yshift=1.4cm] (learn1) {Repair?};
    \node [cblock, below of=learn1, yshift=0.2cm] (repair) {C. Repair offspring};
    \node [block, left of=repair, xshift=-1cm] (evaluate1) {Evaluate offspring};
    \path [line] (begin) -- (initialize);
    \path [line] (initialize) -- (uparchive);
    \path [line] (uparchive) -- (survival);
    \path [line] (survival) -- (terminate);
    \path [line] (terminate) -- node[anchor=west] {Yes} (stop);
    \path [line] (terminate) -- node[anchor=south] {No} (uptarget);
    \path [line] (uptarget) -- (learn);
    \path [line] (mapping) -- (train);
    \path [line] (train) -- (mating);
    \path [line] (mating) -- (learn1);
    \path [line] (repair) -- (evaluate1);
    \path [line] (evaluate1.west) --+(-0.3cm,0) |- (uparchive);
    \path [line] (learn) -- node[anchor=south] {Yes} (mapping);
    \path [line] (learn) -- node[anchor=west] {No} (mating);
    \path [line] (learn1) -- node[anchor=west] {Yes} (repair);
    \path [line] (learn1) -| node[anchor=south] {No} (evaluate1);
\end{tikzpicture}
\caption{EMO/EMaO algorithm with IR2 operator.}\label{process-chart}
\end{figure}

Aligning with the existing IR operator, the new IR2 operator can be used with any suitable EMO/EMaO algorithm, such as NSGA-II \cite{deb2002nsga1}, NSGA-III \cite{nsga3-1,nsga3-2}, MOEA/D \cite{moead}, etc. The main aim towards proposing the IR2 operator is to put emphasis on preserving the diversity in the EMO/EMaO population. The IR2 framework is broadly divided into three steps: (A) \textit{Data Generation} Step, (B) \textit{Training} Step, and (C) \textit{Repair} Step. The overall framework is shown in Figure~\ref{process-chart}, where the three steps above are highlighted in gray. Step~(A) is divided into two parts, i.e., updating target solutions and archive mapping, as explained in Sections~\ref{sec:target-update} and \ref{sec:archive-mapping}, respectively. Steps (B) and (C), i.e., the RF training and repair steps, are detailed in Sections~\ref{sec:training} and \ref{sec:repair}, respectively, followed by their integration with multiple EMO/EMaO algorithms in Section~\ref{sec:overall}. As mentioned earlier in Section~\ref{sec:ANN-repair}, this framework is also guided by two user-defined parameters, i.e., $t_{past}$ and $t_{freq}$, along with a newly introduced parameter $\eta$ to control the greediness of the repair operator. These steps are discussed in more detail in upcoming subsections, followed by the integration algorithm. 

\subsection{Data Generation Step} \label{sec:data-gen-step}
At any generation $t$ of the EMO/EMaO algorithm solving an $M$-objective optimization problem, let the parent and offspring populations, each of size $N$, be referred to as $P_t$ and $Q_t$, respectively. The \textit{archive} $A_t$ (of size $N_A$) consists of the parent population of the $(t-t_{past})^{th}$ generation and offspring populations from $t_{past}$ previous generations. At the same time, a \textit{target-archive} $T_t$ (of size $N$) is also maintained, and is updated in each generation. The archive $A_t$ and target-archive $T_t$ together constitute the \textit{input-output} or \textit{input-target} dataset $D_t$ for the RF training. The process of maintaining the target-archive and of mapping the archive population to these targets for generating the training dataset is discussed in detail in Sections \ref{sec:target-update} and \ref{sec:archive-mapping}, respectively.

\subsubsection{Updating the Target Solutions} \label{sec:target-update}
As mentioned earlier, this learning mechanism ensures that diversity is maintained along with convergence towards the optimal front; i.e., both goals of EMO/EMaO algorithms are pursued together. Towards this realization, a well-distributed target-archive $T_t$ is created and maintained with the help of a uniformly-distributed set of reference points $Z$ (size = $N$, dimensions = $M$). In this study, these reference points were generated using the Das-Dennis method \cite{das}, but they can alternatively be generated by other methods like a recently proposed generic sampling method on an $M$-dimensional unit simplex \cite{unit-simplex-emo}. Since the sizes of $T_t$ and $Z$ are the same, ideally, there should be one target member from $T_t$ associated with each reference point. However, the case of a non-uniform distribution at any generation $t$ has been discussed further.

In order to associate the current population members $P_t$ with the reference points, a normalization procedure is performed in the objective space. Given that $F_k^{(i)}$ represents the $k^{th}$ objective value for the $i^{th}$ individual in $P_t$, $\Bar{F}_k^{(i)}$ represents the $k^{th}$ normalized objective value for the $i^{th}$ individual in $P_t$, and $Z^{ideal}$ and $Z^{nadir}$ represent the ideal and nadir points in $P_t \cup T_t$, the normalization process is given by, 
\begin{eqnarray}
Z_k^{\rm ideal} &=& \min_{i=1}^{N_A} F_k^{(i)}, \label{eq:ideal} \\
Z_k^{\rm nadir} &=& \max_{i=1}^{N_A} F_k^{(i)}, \label{eq:nadir} \\
\bar{F}_k^{(i)} &=& \frac{F_k^{(i)}-Z_k^{\rm ideal}}{Z_k^{\rm nadir} - Z_k^{\rm ideal}}, \quad \forall k=1, \ldots,M
    \label{eq:norm}
\end{eqnarray}

In the normalized objective space, the association of an EMO/EMaO solution with any reference-point can be done in multiple ways, like using Achievement Scalarizing Function (ASF) \cite{wierzbicki}, or Perpendicular Distance Matrix (PDM) \cite{nsga3-1}, or Penalty-based Boundary Intersection (PBI) \cite{moead}. An earlier study \cite{sukrit-coin-ann} discussed the benefits of choosing ASF over PDM for an EMO/EMaO algorithm that does not use any support from reference points, like NSGA-II. For using the proposed operator with other EMO/EMaO algorithms that use a set of reference points, like MOEA/D, NSGA-III, etc., it would be ideal to use the same set of reference points and the same association method as used in the EMO/EMaO algorithm. Hence, we use ASF with a uniform weight vector $w=(1,1,\dots,1)^T$ of size $M \times 1$ for NSGA-II, PDM for NSGA-III and PBI for MOEA/D. 

\begin{figure}[hbt]
    \centering
    \includegraphics[width=.9\linewidth]{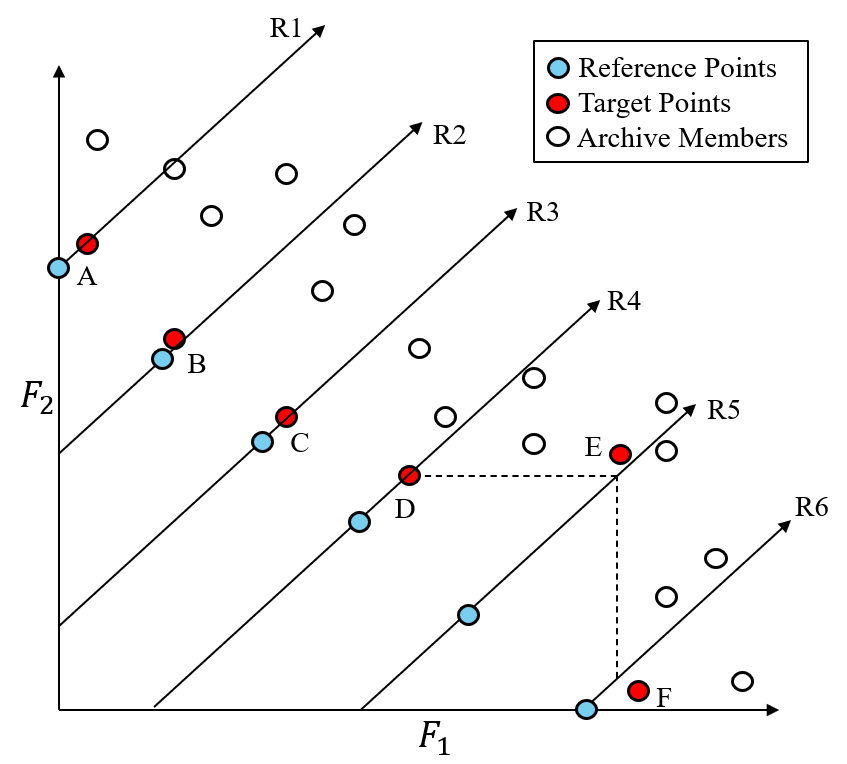}
    \caption{A schematic for identifying target-points using ASF in a two-objective space.}
    \label{fig:target-assoc}
\end{figure}

Target association can be done by two methods, i.e., (i) by identifying the nearest candidate target solution to each reference-vector, and (ii) by identifying the nearest reference-vector to each candidate target solution. While the earlier IR operator uses the former method which ensures that each reference-vector is assigned a target solution (with possibility of duplicates), the proposed IR2 operator uses the latter method which ensures no duplicates in target solutions but may not be able to cover all the reference-vectors. For example, in Figure~\ref{fig:target-assoc}, there is no non-dominated solution associated with the vector R5 (as D dominates E). While the earlier IR operator would have chosen D, based on the smallest ASF value for R5 as the target for R5, the proposed IR2 operator would choose E (a dominated solution, but being close to R5) rather than choosing a solution that belongs to some other reference-vector. If there were no solutions near to the vector R5 at all, there would have been no target assignment for that vector. 

The target archive $T_t$ can have at most $N$ solutions (one for each reference vector). However, as in the above stated example, it may happen that the EMO algorithm does not have any solutions associated with some reference-vectors in the early generations. Due to this, the target archive may have less than $N$ solutions as well for a few generations at the beginning. However, for implementation of IR2 operator, $T_t$ is initialized as an empty array of size $N$, where $T_t^{(i)}$ represents the target solution assigned to the reference point $Z^{(i)}$. The process of updating this target archive from $T_{t-1}$ to $T_t$ using the current parent population $P_t$ is given in Algorithm~\ref{algo:update-target}. Each solution in $P_t$ is evaluated using ASF (or some other metric) to associate it with some reference vector. Let $\alpha$ and $\beta$ be two indices such that the solution $P_t^{(\alpha)}$ gets associated with the reference-vector $Z^{(\beta)}$. Now, if $T_t^{(\beta)}$ (the target assigned to $Z^{(\beta)}$) is empty, that is, no target solution has been associated previously with $Z^{(\beta)}$, then the place is filled with $P_t^{(\alpha)}$. Else, if there is an existing solution at $T_t^{(\beta)}$, then their individual ASF values (with respect to $Z^{(\beta)}$) are compared to determine if the replacement is needed.

\begin{algorithm}[hbt]
\caption{$T_t =$ Update\_Target ($P_t$, $T_{t-1}$, $Z$)}\label{algo:update-target}
\begin{algorithmic}[1]
\Require Reference points $Z$, Target archive $T_t$, Parent Population $P_t = \left(\boldx^{(1)},\ldots,\boldx^{(N)}\right)$
\State $F_T \gets \text{Objective function values in } T_{t-1}$
\State $F_P \gets \text{Objective function values in } P_t$
\State Compute $Z^{\rm ideal}$ \& $Z^{\rm nadir}$ from $F_P$ \hfill \texttt{\%Eqs.~\ref{eq:ideal}-\ref{eq:nadir}}
\State $\bar{F_P} \gets \text{Normalize } F_P \text{ using } Z^{\rm ideal} \text{ and } Z^{\rm nadir}$ \hfill \texttt{\%Eq.~\ref{eq:norm}}
\State $\bar{F_T} \gets \text{Normalize } F_T \text{ using } Z^{\rm ideal} \text{ and } Z^{\rm nadir}$ \hfill \texttt{\%Eq.~\ref{eq:norm}}
\For{$i$ = 1 to $N$} \hfill \texttt{\% for each pt. in $F_P$}
\For{$j$ = 1 to $N$} \hfill \texttt{\% for each pt. of $Z$}
\State $ASF_j \gets \max_{k=1}^M \left(\bar{F}_{P,k}^{(i)}-Z_k^{(j)}\right)$
\EndFor
\State $[Index,ASF_P] \gets \min_{j=1}^{N} ASF_j$  
\If{$T_t^{(Index)}$ is empty}
\State $T_t^{(Index)} \gets P_t^{(i)}$
\Else
\State $ASF_T \gets \max_{k=1}^M \left(\bar{F}_{T,k}^{(Index)}-Z_k^{(Index)}\right)$
\If{$ASF_P < ASF_T $}
\State $T_t^{(Index)} \gets P_t^{(i)}$
\EndIf
\EndIf
\EndFor
\State \Return Updated target-archive $T_t$
\end{algorithmic}
\end{algorithm}


\subsubsection{Archive Mapping} \label{sec:archive-mapping}

The goal is to map the solutions in archive $A_t$ to those in the target archive $T_t$ to generate a training dataset that is then fed to train an ML model or a multi-target regressor.

This goal is realized by associating a solution in $A_t$ with one of the reference points, say $Z^{(\beta)}$, for which a corresponding target solution $T_t^{(\beta)}$ already exists. However, it may be noted that to normalize the archive solutions, the $Z^{ideal}$ and $Z^{nadir}$ are calculated separately for $A_t$, and may be different from the ideal and nadir points of the target archive $T_t$. This helps in ensuring that if diversity has been lost in the past few generations, it can again be regained (partially) by this learning-repair process. Following this, each member in archive $A_t$ is associated with one reference point---say $A_t^{(\alpha)}$ is associated with $Z^{(\beta)}$. Then, $A_t^{(\alpha)}$ goes to the $Input$ and $T_t^{(\beta)}$ goes to the $Output$, which then forms one $Input$-$Output$ sample for the training dataset $D_t$. This process is shown in Algorithm~\ref{algo:archive-mapping}, which returns the dataset used for ML training.


\begin{algorithm}[hbt]
\caption{$D_t =$ Archive\_Mapping ($A_t,T_t,Z$)}\label{algo:archive-mapping}
\begin{algorithmic}[1]
\Require Archive $A_t$, Target-archive $T_t$, Ref.-points $Z$.
\State $N_A \gets \text{Size of Archive } A_t$
\State $F_A \gets \text{Objective function values in } A_t$
\State Compute $Z^{\rm ideal}$ \& $Z^{\rm nadir}$ from $F_A$ \hfill \texttt{\%Eqs.~\ref{eq:ideal}-\ref{eq:nadir}}
\State $\bar{F_A} \gets \text{Normalize } F_A \text{ using } Z^{\rm ideal} \text{ and } Z^{\rm nadir}$ \hfill \texttt{\%Eq.~\ref{eq:norm}}
\State $Input \gets \emptyset$
\State $Output \gets \emptyset$
\For{$i$ = 1 to $N_A$} \hfill \texttt{\% for each pt. in $A_t$}
\For{$j$ = 1 to $N$} \hfill \texttt{\% for each pt. in $Z$}
\State $ASF_j \gets \max_{k=1}^M \left(\bar{F}_{A,k}^{(i)}-Z_k^{(j)}\right)$
\EndFor
\State $Index \gets \text{arg}\_\min_{j=1}^{N} ASF_j$ 
\State $Input \gets Input \cup A_t^{(i)}$
\State $Output \gets Output \cup T_t^{(Index)}$
\EndFor
\State $\boldx_{Input} \gets \text{Variable vectors in } Input$
\State $\boldx_{Output} \gets \text{Variable vectors in } Output$
\State $D_t \gets [\boldx_{Input}, \boldx_{Output}]$
\State \Return Training Dataset $D_{t}$
\end{algorithmic}
\end{algorithm}

\subsection{Training Step} \label{sec:training}

After the training dataset $D_t$ is generated, the ML model is trained to capture the $Input$-$Output$ relationships of the decision variable vectors. The training process is presented in Algorithm~\ref{algo:training}, which mainly consists of two steps, i.e., the \textit{dynamic normalization} (discussed in our earlier work \cite{sukrit-coin-ann}) of training-dataset $D_t$ (a pre-training step) and the training of the ML model itself. 



\begin{algorithm}[hbt]
\caption{$\left(\texttt{Predict\!()}\!,\![\!\boldx^{\text{min}}\!,\!\boldx^{\text{max}}]\!\right)\!=\!$Training($D_t,[\boldx^l,\boldx^u]$)}\label{algo:training}
\begin{algorithmic}[1]
\Require Training dataset $D_t$, Lower \& upper variable-bounds specified in the problem $\boldx^l$ and $\boldx^u$.
\State $\boldx^{l,t} \gets \text{Min. variable vector in $Input$-$Output$ data } D_t$
\State $\boldx^{u,t} \gets \text{Max. variable vector in $Input$-$Output$ data } D_t$
\For{$k$ = 1 to $n_{var}$} \hfill \texttt{\% No. of variables}
\State $x_{k}^{\text{min}} = 0.5(x_k^{l,t} + x_{k}^{l})$
\State $x_k^{\text{max}} = 0.5(x_k^{u,t} + x_k^{u})$
\EndFor
\State Normalize $D_t$ using $\boldx^{\text{min}}$ and $\boldx^{\text{max}}$ as bounds
\State Train the ML model using $D_t$ to create \texttt{Predict()}
\State \Return \texttt{Predict()}, Bounds [$\boldx^{\text{min}}, \boldx^{\text{max}}$]
\end{algorithmic}
\end{algorithm}

\subsection{Repair Step} \label{sec:repair}

The trained ML model is then used to repair the offspring solutions $Q_t$ obtained after genetic recombination and mutation operations at generation $t$. This repair step consists of 4 different processes, i.e., 

\begin{enumerate}
    \item Selecting 50\% of offspring randomly and repairing them using the \texttt{Predict()} function obtained after training the ML training.
    \item \textit{Near-bound Restoration --} Any variable near one of its bounds (before Repair) is restored to its original value. 
    \item \textit{Boundary Repair --} Any variable that lies outside its respective permissible bounds [$\boldx_k^l, \boldx_k^u$] is fixed to an inner value based in Inverse Parabolic Spread Distribution \cite{boundary-PSO}.
    \item \textit{Enhancement} of the repaired solutions. 
\end{enumerate}

Process-4 is newly introduced, as explained further in the next subsection.

\subsubsection*{Enhancement of Repaired solutions}
A new parameter governs enhancement of the repaired solutions, hereafter referred to as $\eta$. At any intermediate generation, the ML model would learn the $Input$-$Output$ relationships, which could have been learned from the best solutions belonging to the already explored search space. However, the global optima may lie in regions of the search space yet to be explored, implying that the relationships learned until now are prematurely restrictive. Given that, for an offspring $X$ and a repaired offspring $Y$, a greedy algorithm would exploit this possibility by further extrapolating the solution in the decision space. This exploitation is done using parameter $\eta$ by defining the enhanced solution $Y'$ as,
\begin{equation}
    Y' = X + \eta(Y-X),
\end{equation}
where $\eta=1$ is equivalent to no enhancement, and any value of $\eta>1$ will potentially enhance the repaired solutions. However, too much greediness can be disadvantageous as well, a study of which is done in \ref{sec:parameter-settings}. In the future, this parameter can be made adaptive as the EMO/EMaO algorithm progresses. However, in this paper it is kept fixed to the value defined while initializing the EMO/EMaO algorithm. The complete repair process is presented in Algorithm~\ref{algo:repair}.

\begin{algorithm}[hbt]
\caption{$Q_t\!=$Repair($Q_t$,$\eta$,[$\boldx^{\text{min}},\boldx^{\text{max}}$],[$\boldx^l,\boldx^u$],\texttt{Predict})}\label{algo:repair}
\begin{algorithmic}[1]
\Require Offspring $Q_t$, Enhancement factor $\eta$, Bounds from Algorithm~\ref{algo:training} [$\boldx^{\text{min}},\boldx^{\text{max}}$], Variable bounds in problem [$\boldx^l,\boldx^u$], \texttt{Predict()} function.
\State $I \gets \text{Randomly selected 50\% }Q_t$
\State $range_k \gets \left(x_k^u-x_k^l\right)$, \ $k = 1,\ldots,n_{var}$
\For{$i=1$ to $N/2$}
\State $temp \gets \text{Normalize $I^{(i)}$ using } \boldx^{\min} \text{ and } \boldx^{\max}$
\State $R^{(i)} \gets \texttt{Predict}(temp)$
\State $R^{(i)} \gets \text{Denormalize $R^{(i)}$ using } \boldx^{\min} \text{ and } \boldx^{\max}$
\State $R^{(i)} \gets I^{(i)} + \eta \left(R^{(i)}-I^{(i)}\right)$
\For{k = 1 to $n_{var}$}
\State $vicinity \gets \min\left(|I^{(i)}_k\!-\!x_k^{\min}|,|x_k^{\max}\!-\!I^{(i)}_k|\right)$
\If {$vicinity\leq 0.01*range_k$}
\State $R^{(i)}_k \gets I^{(i)}_k$
\EndIf
\State Replace $I^{(i)}$ by $R^{(i)}$ in $Q_t$
\EndFor
\State $\text{Boundary Repair on } I^{(i)}$
\EndFor
\State \Return Repaired offspring $Q_t$
\end{algorithmic}
\end{algorithm}

\subsection{Learning Termination} \label{sec:termination}
The kind of learning-based repair operator described above can help gain faster convergence by improving the fitness of the offspring generated by crossover and mutation operators \cite{sukrit-ann-repair-gecco,sukrit-coin-ann}. However, the effect of repairs is bound to decrease in later generations when the population is nearly converged, as compared to earlier generations. Hence, a termination criterion is incorporated, which monitors the improvement in solutions and adaptively turns off the learning when the improvement becomes relatively smaller. Since convergence is the only criterion for termination here, the metric used is \textit{GD} (Generational Distance), which is evaluated at generation $t$ for $P_d$ (where $d=t-t_{past}$) with the current generation's population as the reference set. This running metric will be denoted by $G$, given as,
\begin{equation}
    G_{t}(P_d) = \frac{1}{|P_d|}\sqrt{\Sigma_{i=1}^{|P_d|} d_{i}^{2}},
    \label{eq:g-metric}
\end{equation}
where $d_i$ represents the euclidean distance (in $F$-space) from a member in $P_d$ to its nearest population member in $P_t$. This parameter $G$ is monitored at every generation after the repair operator is activated. If the value of $G_t$ drops below $g_{th}$\% 
of the maximum value of $G$ until then, the learning and repair steps are switched off, and the EMO/EMaO algorithm resumes its original operation. However, if $G_t$ becomes higher than this threshold at some later generation, then the learning and repair operations resume. This is a probable scenario in problems where the base EMO/EMaO algorithm usually gets stuck in a local optimum. The effect of this threshold parameter for termination is discussed in Section~\ref{sec:termination-discussion}.


\subsection{Integration with EMO/EMaO Algorithms} \label{sec:overall}

\subsubsection{NSGA-II and NSGA-III}

The integration of the proposed IR2 operator with NSGA-II and NSGA-III is presented in Algorithm~\ref{algo:overall}. The pseudo-code represents an intermediate generation of the optimization algorithm along with IR2-related steps. First, the target archive $T_t$ is updated using the parent population $P_t$. Lines 2--7 check whether to use the IR2 operator as per parameter $t_{freq}$ and the termination criterion given in Section~\ref{sec:termination}. If repair is to be done, the $flag$ is marked \textit{True}, else \textit{False}. In the learning step, mapping is done using the \texttt{Archive\_Mapping} function (Algorithm~\ref{algo:archive-mapping}) and then training is done using the \texttt{Training} function (Algorithm~\ref{algo:training}), as given in lines 8--10. Following this, mating is done using genetic operators (recombination and mutation, line 11) along with the repair process using the \texttt{Repair} function (Algorithm~\ref{algo:repair}). The repair process (line 13) is done before the offspring fitness evaluation step (line 14) so that the offspring are evaluated only once, in line with the EMO algorithm. Line 15 shows the update of archive $A_t$ as explained initially in Section~\ref{sec:data-gen-step}, followed by survival selection in line 16.

\begin{algorithm}[hbt]
\caption{Generation $t$ of NSGA-II and NSGA-III with IR2 operator.}\label{algo:overall}
\begin{algorithmic}[1]
\Require Reference Set $Z$, Parent population $P_t$, Archive $A_t$, Previous Target-archive $T_{t-1}$, Collection $t_{past}$, Frequency $t_{freq}$, Enhancement Factor $\eta$, Variable bounds defined in problem [$\boldx^l,\boldx^u$].
\State $T_{t} \gets \mbox{Update\_Target}(P_t,T_{t-1},Z)$
\State $count \gets \mbox{rem}(t, t_{\rm freq})$
\State $G \gets G \cup G_t$ \hfill \texttt{\% Equation \ref{eq:g-metric}}
\If {$G_t>g_{th} \% \text{ of } \text{max}(G) \text{ \& } count=0$}
\State $flag \gets$ True
\Else 
\State $flag \gets$ False
\EndIf
\If{$flag=$True}
\State $D_t \gets \mbox{Archive\_Mapping}(A_t,T_t,Z)$
\State $\left(\texttt{Predict}, [\boldx^{\text{min}}, \boldx^{\text{max}}]\right) \gets \mbox{Training}(D_t,[\boldx^l, \boldx^u])$
\EndIf
\State $Q_t \gets \text{Recombination + Mutation}(P_t)$
\If{$flag=\text{True}$}
\State $Q_t \gets \mbox{Repair}$($Q_t$,$\eta$,[$\boldx^{\text{min}}, \boldx^{\text{max}}$],[$\boldx^l, \boldx^u$],\texttt{Predict})
\EndIf
\State $\text{Evaluate } Q_t$
\State $A_{t+1} \gets \left(A_t \cup Q_t \cup P_{t+1-t_{\rm past}} \right)\backslash [P_{t-t_{\rm past}} \cup Q_{t-t_{\rm past}}]$
\State $P_{t+1} \gets \text{Survival selection of NSGA-II/III with } P_t \cup Q_t$
\State \Return Next Parent Population $P_{t+1}$, Next Archive $A_{t+1}$, Target-archive $T_{t}$
\end{algorithmic}
\end{algorithm}

\subsubsection{MOEA/D}

The integration of the proposed IR2 operator with MOEA/D is presented in Algorithm~\ref{algo:overall-moead}. The pseudo-code represents an intermediate generation of the optimization algorithm along with IR2 related steps. First, the target archive $T_t$ is updated using the parent population $P_t$. Lines 2--7 check whether to use the IR2 operator as per the parameter $t_{freq}$ and the termination criterion given in Section~\ref{sec:termination}. If repair is to be done, the $flag$ is marked \textit{True}, else \textit{False}. In the learning step, mapping is done using the \texttt{Archive\_Mapping} function (Algorithm~\ref{algo:archive-mapping}) and then training is done using \texttt{Training} function (Algorithm~\ref{algo:training}), as given in lines 8--10. Following this, 50\% of solutions are randomly selected (line 12), and their indices are stored in $I$. Lines 14--15 represent the selection and mating process---i.e., whether the parents are selected from the neighborhood or the entire population---followed by creation of a single offspring. If the index of this offspring was selected in $I$, it is repaired using the \texttt{Repair} function (Algorithm \ref{algo:repair}). The offspring are then evaluated, and neighborhood solutions are updated according to the MOEA/D algorithm. In the end, line 20 shows the update of Archive $A_t$.
\begin{algorithm}[hbt]
\caption{Generation $t$ of MOEA/D with IR2 operator.}\label{algo:overall-moead}
\begin{algorithmic}[1]
\Require Reference Set $Z$, Parent population $P_t$, Neighborhood solutions $B$, Neighborhood size $N_S$, Population Size $N$, Archive $A_t$, Previous Target-archive $T_{t-1}$, Collection $t_{past}$, Frequency $t_{freq}$, Enhancement Factor $\eta$, Variable bounds defined in problem [$\boldx^l,\boldx^u$].
\State $T_{t} \gets \mbox{Update\_Target}(P_t,T_{t-1},Z)$
\State $count \gets \mbox{rem}(t, t_{\rm freq})$
\State $G \gets G \cup G_t$ \hfill \texttt{\% Equation \ref{eq:g-metric}}
\If {$G_t>g_{th} \% \text{ of } \text{max}(G) \text{ \& } count=0$}
\State $flag \gets$ True
\Else 
\State $flag \gets$ False
\EndIf
\If{$flag=$True}
\State $D_t \gets \mbox{Archive\_Mapping}(A_t,T_t,Z)$
\State $\left(\texttt{Predict}, [\boldx^{\text{min}},\boldx^{\text{max}}]\right) \gets \mbox{Training}(D_t,[\boldx^l,\boldx^u])$
\EndIf
\State $P_{t+1} \gets P_t$
\State $I \gets$ 50\% Randomly selected indices from $N$
\For{$i=1$ to $N$}
\State $j,k \gets$ Two selected indices from $B(i)$ or $P_t$
\State $Q_t^{(i)} \gets \text{Recombination + Mutation}(P_t^{(j)},P_t^{(k)})$
\If{$flag=$True and $i \in I$}
\State $Q_t^{(i)} \gets \mbox{Repair}$($Q_t^{(i)}$,$\eta$,[$\boldx^{\text{min}},\boldx^{\text{max}}$],\texttt{Predict})
\EndIf
\State $\text{Evaluate } Q_t^{(i)}$
\State Update the neighborhood solutions $B(i)$ in $P_{t+1}$
\EndFor
\State $A_{t+1} \gets \left(A_t \cup Q_t \cup P_{t+1-t_{\rm past}} \right)\backslash [P_{t-t_{\rm past}} \cup Q_{t-t_{\rm past}}]$
\State \Return Next Parent Population $P_{t+1}$, Next Archive $A_{t+1}$, Target-archive $T_{t}$
\end{algorithmic}
\end{algorithm}

Integration with other EMO/EMaO algorithms can also be suitably achieved. 

\subsection{Differences with earlier IR operator} \label{sec:differences}

Apart from changing the associated learning technique from ANN to RF, there are certain other differences between the core of the IR and IR2 operators with a primary focus on preserving diversity in population and some other issues, as enumerated below.

\begin{enumerate}
    \item \textit{Archive Members --} The IR operator has an archive $A_t$ that contains solutions from $t_{past}$ previous generations. Since not all parents $P_t$ are replaced by offsprings $Q_t$ in the survival selection, this archive will have multiple copies of same solution, more so in later generations where only a small fraction of offsprings might get selected. This results in duplicates in the training dataset $D_t$, undermining its generality. To address this, the archive $A_t$ in the IR2 operator stores offspring solutions from $t_{past}$ previous generations and one parent population $P_{t-t_{past}}$. This new archive covers all the solutions in the earlier definition of archive and has possibly no duplicates, advocating for a better learning process. 
    \item \textit{Target Archive --} Unlike the IR operator that chooses the target solutions from the non-dominated solutions in the current population $P_t$, the IR2 operator maintains a separate target-archive of the $N$ best solutions found across each of the $N$ reference vectors. The target-archive $T_t$ is updated at every generation, from the solutions in $P_t$ if found better than any of the existing solutions in $T_t$. This ensures that each part of the obtained front contributes to the training dataset, even though some part may be dominated from some other target solutions. With this modification, the requirement of 50\% of $P_t$ to be non-dominated for learning is removed from the IR2 operator, which was there earlier in the IR operator. 
    \item \textit{Target Association --} In the IR operator, each reference vector is supplied with it's nearest non-dominated solution as the target, even though that solution may belong in the region of some other vector. In case the EMO algorithm has no non-dominated solution for a particular vector, the IR operator will map the archived solutions on that vector to the assigned target solution (which would belong to some other reference vector). This may create some noise in the dataset and may contribute in further widening the gap in the front. Towards this effect, the IR2 operator checks the association of a current solution with a reference vector, and if found better, then assigns it to that particular vector only. This would mean that some reference-vectors may not have a target solution in the initial generations, but then they simply would not be considered for training unless they have some representative solutions. 
    \item \textit{Boundary repair --} In the IR operator, if any variable goes out of its respective bounds during the repair step, that variable is fixed at the respective boundary value. However, in case of IR2, the boundary repair value is determined using an inverse parabolic spread distribution \cite{boundary-PSO}. This would be beneficial since fixing a variable to its boundary in multiple solutions may cause inefficiencies for the genetic mating operators. 
    \item \textit{Enhancement of repaired offsprings --} This is a new proposition in the IR2 operator, which did not exist in the IR operator earlier. This enhancement contributes directly towards greediness of the repair operator, an effect of which is visualized through a small parametric study in Section~\ref{sec:parameter-settings}. 
\end{enumerate}

It is evident that points 2--4 mentioned above focus on maintaining the diversity in either $F$- or $X$-space, which advocates for the main aim of designing this modified IR2 operator. These inclusions make IR2 a more robust operator to be used for a wide range of optimization problems. The efficacy of this operator is discussed in the following sections. 

\section{Results with IR2 Operator} \label{sec:results}
In this section, the performance of the IR2 operator is analyzed for three EMO/EMaO algorithms, i.e., NSGA-II \cite{deb2002nsga1}, NSGA-III \cite{nsga3-1,nsga3-2} and MOEA/D \cite{moead}. These algorithms, integrated with the IR2 operator and RF as the ML technique, are hereafter called NSGA-II-IR2-RF, NSGA-III-IR2-RF and MOEA/D-IR2-RF. The main aim of the proposed IR2 operator is to assist the underlying EMO/EMaO algorithm to converge faster, while ensuring that the diversity in $F$-space is maintained. In other words, the IR2 operator is expected to achieve a result in, say, $t$ iterations, for which the EMO/EMaO algorithm without repair would have taken $t+\Delta t$ iterations. Since the IR2 operator, and even the IR operator, do not require any extra function evaluations, the $\Delta t$ iterations and corresponding function evaluations can be seen to be a direct saving. This benefit is bound to decrease as the optimization progresses and the population reaches the vicinity of the PO front. However, in real-world problems, the number of allowed function evaluations are usually limited because they are computationally expensive. Hence, we try to assess the efficacy of the proposed IR2 operator on several 2--5-objective benchmark test problems, analyzing the results at intermediate generations. 

For this analysis, we primarily use the hypervolume (HV) metric, since it is indicative of both convergence and diversity of the solution set. At any intermediate generation $t$ for a particular test instance, the median hypervolume (from multiple independent seed runs) is recorded for both algorithms, an EMO/EMaO algorithm with and without the IR operator. Let us call these HV values as $H_{R}^t$ and $H^t$, respectively. Since these values are different, let us say that the EMO/EMaO algorithm without repair takes $t+\Delta t$ iterations to reach to an equivalent HV, i.e., $H_R^t = H^{t+\Delta t}$. $\Delta t$ may be negative if the IR operator makes the EMO/EMaO algorithm perform worse than the original algorithm. The percentage savings ($S$) in the function evaluations can be calculated as,
\begin{equation} \label{eq:S-metric}
    S = \frac{\Delta t}{t} \times 100 \%
\end{equation}

The $S$-metric is a direct indicator of the benefits of using the proposed IR2 or earlier IR operator. The comparison on different problem instances is done only with the same underlying algorithm, since we propose this as a generic learning-based repair operator which can be used with any other learning algorithm (than RF) or any other EMO/EMaO algorithm. Apart from this, the final generation results are often similar with/without the IR/IR2 operator, since both algorithms approach the Pareto front. However, the end results are also better with the IR/IR2 operator in some instances where the genetic operators are simply unable to converge.


For performance comparison, hypervolume (HV) has been used to quantify the measure of convergence and diversity. For some problems, Generational Distance (GD) and Inverted Generational Distance (IGD) has also been computed. The HV-computation has been done with reference point as $[P]_{(1 \times M)}$ for $M$-objective problems, where $P=\frac{N}{N-1}$ ($N$ is the population size). The GD and IGD computation has been done with a Pareto front of 500 equally space data points for 2-objective problems. Also, $p$-values have been calculated using the Wilcoxon Rank-Sum test to supplement the visually apparent differences with statistical reasoning. For WFG problems, where the PO front does not lie between 0 and 1, the $i^\text{th}$ objective is normalized with bounds $[0,2*i]$ before hypervolume is computed. 

\subsection{Test Suite} \label{sec:test-suite}
Several two to five-objective optimization problems have been used in this paper, in order to establish the improved performance of an EMO/EMaO algorithm using the IR2 operator as compared to the base EMO/EMaO algorithm. Before starting with the extensive results, a small parametric study for the enhancement parameter $\eta$ is presented in Section~\ref{sec:parameter-settings}, for which two problems L1 and L2 have been taken from \cite{sukrit-coin-ann}. The dependency on the learning termination threshold is discussed in Section~\ref{sec:termination-discussion}.


For demonstrating the efficacy of the proposed IR operator, several benchmark test suites have been used, namely ZDT \cite{ZDT}, WFG \cite{WFG} and DTLZ \cite{DTLZ}. In the ZDT problems, their respective $g(X)$-functions have been modified to have PO solutions at $x_k=0.5$ $\forall$ $k=2,\dots,30$. Further, WFG problems are solved with $M=3,4$ and total 24 variables. For these problems, the position-related variables $k$ are fixed as $k=2 \times (M-1)$. The difficulty of these problems is scalable and parameter dependent, and these values are generally taken as the defaults from \cite{WFG}. In some instances, these parameters are modified to increase the difficulty (discussed in Section~\ref{sec:discussion}). For the DTLZ problems, the number of objectives is varied as $M=3,4,5$ while the number of variables is fixed at 15. 

\subsection{Parameter Settings} \label{sec:parameter-settings}

The base EMO/EMaO algorithms deployed with the RF-assisted IR2 operator are NSGA-II (for 2-objective problems), and NSGA-III and MOEA/D (for problems with more than two objectives). NSGA-III is chosen over NSGA-II for three-objective problems since studies suggest that NSGA-III outperforms NSGA-II on these problems \cite{NSGA-II-III-comparison}. For all problems, SBX crossover ($p_c=0.9$ and $\eta_c=10$), and polynomial mutation ($p_m=0.1$ and $\eta_m=1/n_{var}$) is used. The population size $N$ is kept different, varying with the number of objectives $M$ as given in Table~\ref{tab:population-size}, along with number of gaps $p$ for the Das-Dennis method to generate exactly $N$ reference-points.

\begin{table}[hbt]
    \caption{Population Size $N$ and gaps for Das-Dennis method $p$ for problems with $M$ objectives.}
    \label{tab:population-size}
    \centering
    \begin{tabular}{|c|c|c|c|c|} \hline
        $M$ & 2 & 3 & 4 & 5\\ \hline
        $N$ & 100 & 105 & 286 & 495 \\ \hline
        $p$ & 99 & 13 & 10 & 8 \\ \hline
    \end{tabular}
\end{table}

From \cite{sukrit-coin-ann}, the archive collection parameter $t_{past}$ and the frequency of repairs $t_{freq}$ are both fixed at 5. The parameters used for RF training have been discussed earlier in Section~\ref{sec:IR-RF}.

\begin{table*}[]
\caption{Effect of repair enhancement parameter $\eta$ on two difficult problems L1 and L2.}
\label{tab:param-n}
\centering
\begin{tabular}{|l|c|c|c|c|c|c|c|c|}
\hline
\multicolumn{1}{|c|}{\multirow{2}{*}{Problem}} & \multicolumn{2}{c|}{$\eta$=1.0} & \multicolumn{2}{c|}{$\eta$=1.1}     & \multicolumn{2}{c|}{$\eta$=1.2} & \multicolumn{2}{c|}{$\eta$=1.3} \\ \cline{2-9} 
\multicolumn{1}{|c|}{}                         & median-HV    & $p$-value     & median-HV         & $p$-value    & median-HV    & $p$-value     & median-HV    & $p$-value     \\ \hline
L1                                             & 0.578317     & 5.11E-06    & \textbf{0.586048} & -- & 0.57954      & 5.28E-03   & 0.574252     & 6.47E-05    \\ 
L2                                             & 0.662975     & 1.40E-04   & \textbf{0.665669} & -- & 0.664926     & 2.56E-04   & 0.664079     & 1.41E-06    \\ \hline
\end{tabular}
\end{table*}

\begin{figure}[hbt]
    \centering
\begin{subfigure}[b]{0.9\linewidth}
\centering    \includegraphics[width=\linewidth]{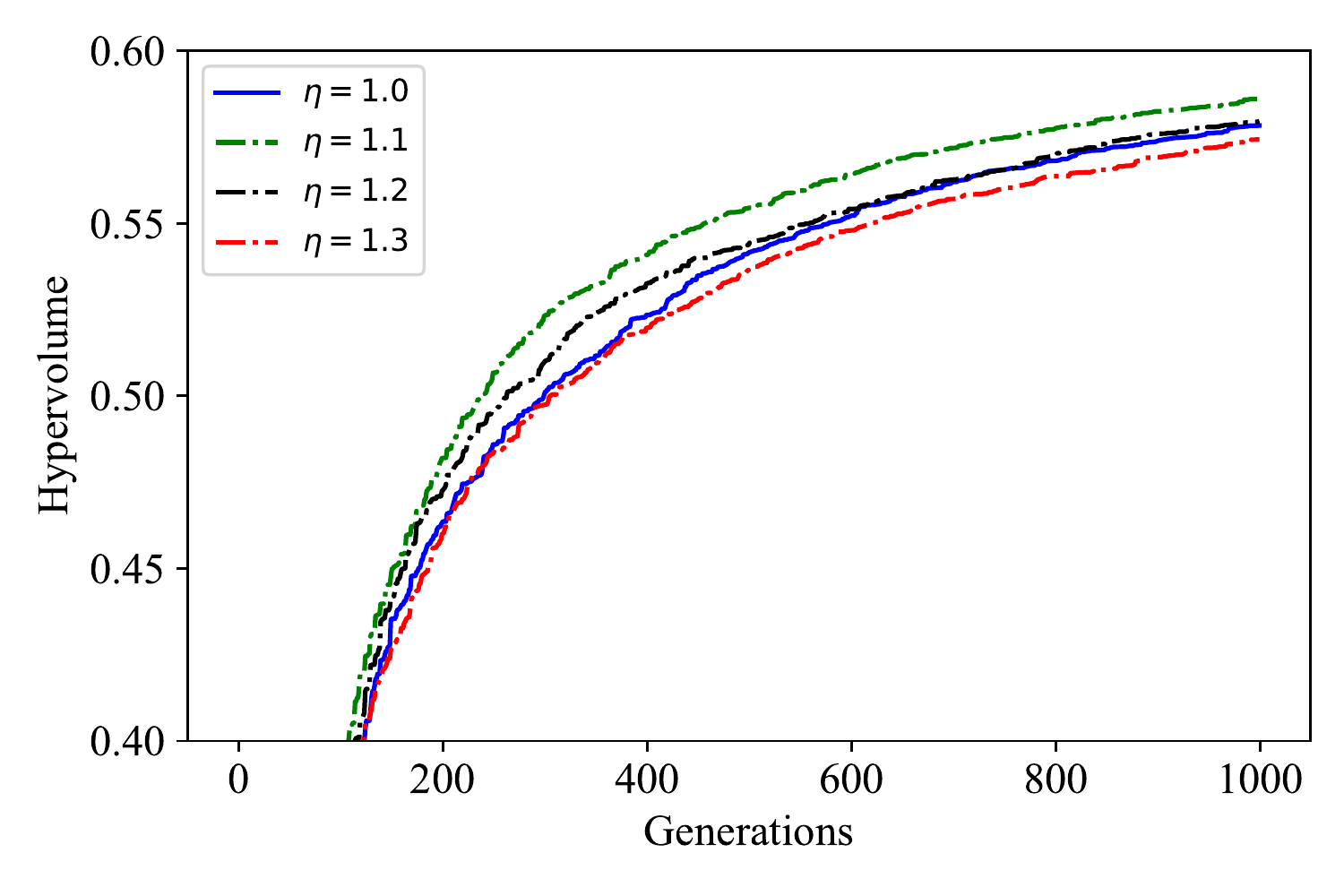}
    \caption{Problem L1.}
    \label{fig:param-n-L1}
    \end{subfigure}\\
\begin{subfigure}[b]{0.9\linewidth}
    \centering
    \includegraphics[width=\linewidth]{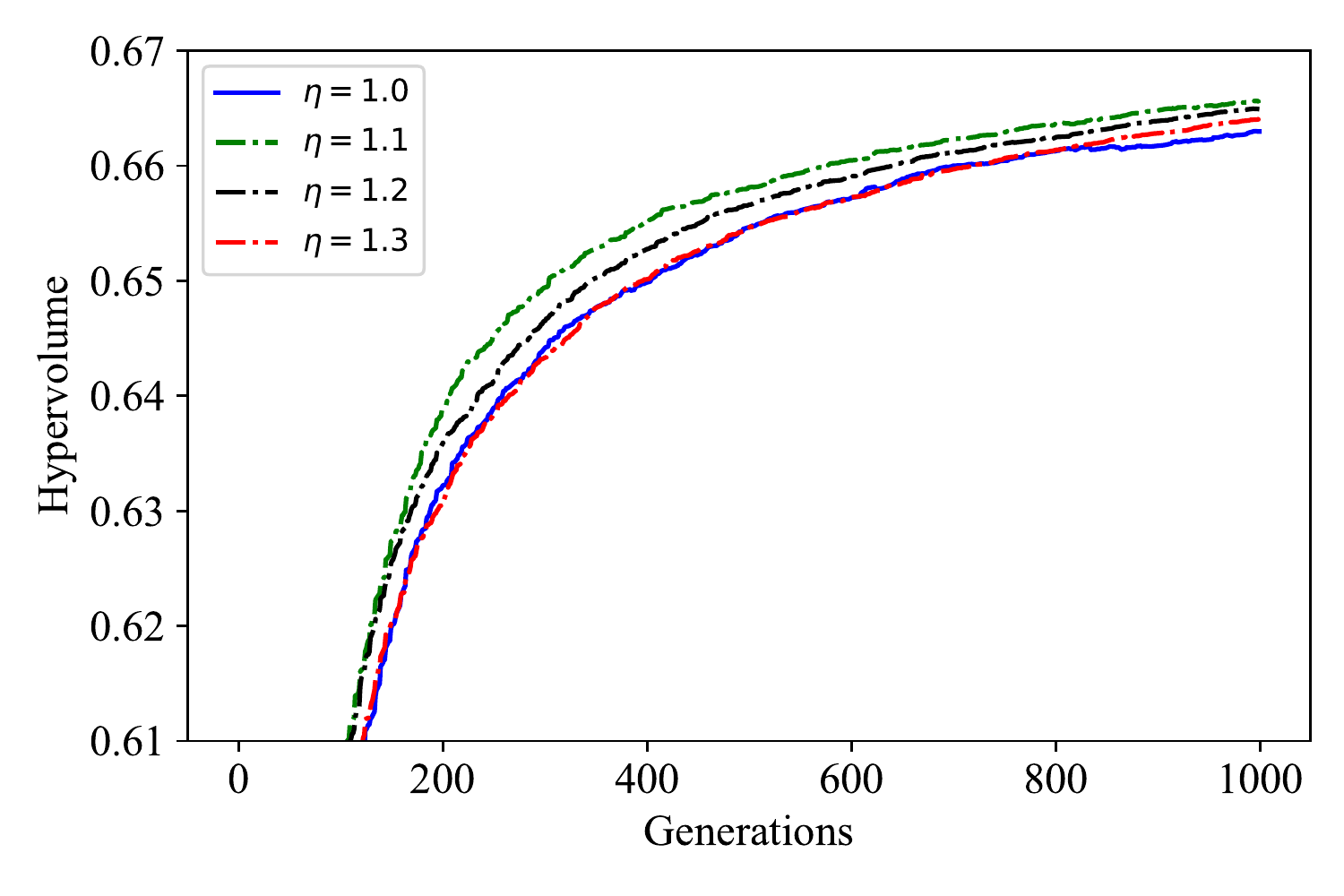}
    \caption{Problem L2.}
    \label{fig:param-n-L2}
\end{subfigure}
\caption{Median HV for different $\eta$ on problems L1 and L2.}
\label{fig:param-n}
\end{figure}

We investigate the behaviour of the IR2-RF operator with respect to the newly introduced enhancement operator $\eta$. For the parametric study, we explore the values of $\eta$ as 1.0 (i.e., no enhancement), 1.1, 1.2 and 1.3. The median hypervolumes from 16 independent runs over 1,000 generations are shown in Figure~\ref{fig:param-n} for two problems, L1 and L2 \cite{sukrit-coin-ann}. It can be observed that there is a consistent advantage at almost all generations with $\eta=1.1$. Table~\ref{tab:param-n} also shows the median HV values at the final generation and their respective $p$-values. 
However, even though $\eta=1.1$ produces the best results, it is noteworthy that solutions with $\eta>1$ are better than no enhancement ($\eta=1$) at all.
To conclude, this preliminary parametric study motivates us to use $\eta=1.1$ for all subsequent experiments in this paper.

\subsection{ZDT problems}

First, we discuss and analyze the performance of the proposed IR2 operator (NSGA-II-IR2-RF) as compared to the base algorithm NSGA-II and NSGA-II-IR-RF (discussed earlier in Section~\ref{sec:IR-RF}). For a fair comparison, RF is chosen as the learning method for both IR and IR2 operators. Their median hypervolume values at generation 100 are shown in Table~\ref{tab:ann-rf}. It is evident from the table that NSGA-II-IR2-RF performs either better, or, at par in all test instances. The better performance of NSGA-II-IR2-RF over NSGA-II-IR-RF implies that the modified IR operator is more effective, considering that the learning method (RF) is the same in both algorithms.

Further, the same ZDT problems are used to compare MOEA/D and MOEA/D-IR2-RF on two-objective problems. At generation 50, the HV, IGD and GD values are shown in Table~\ref{tab:moead-zdt} along with their respective $p$-values. MOEA/D-IR2-RF either outperforms, or, is at par with MOEA/D on all problems, except ZDT4. In ZDT4, the GD for MOEA/D is better but the results are not statistically different. 
\begin{table*}[hbt]
\caption{Hypervolume values for two-objective problems, comparing NSGA-II without repair, and with RF-assisted IR and IR2 operators. Best performing framework is marked in bold.}
\label{tab:ann-rf}
\centering
\begin{tabular}{|c|c|c|c|c|c|c|c|}
\hline
\multirow{2}{*}{Problem} & \multirow{2}{*}{Generation} & \multicolumn{2}{c|}{NSGA-II} &  \multicolumn{2}{c|}{NSGA-II-IR-RF} & \multicolumn{2}{c|}{NSGA-II-IR2-RF} \\ \cline{3-8}
                         &                             & median-HV     & p-value       & median-HV       & p-value       & median-HV           & p-value  \\ \hline
ZDT1                     & 100                          & 0.675528       & 1.33E-11    & 0.678876         & 1.63E-02      & \textbf{0.679119}   & -        \\
ZDT2                     & 100                          & 0.340607      & 1.33E-11    & 0.344396        & 1.97E-07      & \textbf{0.345401}   & -        \\
ZDT3                     & 100                          & 0.532092      & 1.33E-11    & \textit{0.534475}        & 1.26E-01      & \textbf{0.534578}   & -        \\
ZDT4                     & 100                          & \textit{0.673986}      & 1.06E-01    & \textit{0.675198}        & 3.49E-01      & \textbf{0.676978}   & -        \\
ZDT6                     & 100                          & 0.159995      & 6.34E-09    & 0.169844        & 5.19E-07      & \textbf{0.230231}   & -      \\ \hline
\end{tabular}
\end{table*}

\begin{table*}[hbt]
\caption{Performance metrics for two-objective problems, comparing MOEA/D without repair, and with repair (MOEA/D-IR2-RF). Best performing framework is marked in bold.}
\resizebox{\linewidth}{!}{
\label{tab:moead-zdt}
\centering
\begin{tabular}{|c|c|ccc|ccc|ccc|}
\hline
\multirow{2}{*}{Problem} & \multirow{2}{*}{Generation} & \multicolumn{3}{c|}{median-HV}                                   & \multicolumn{3}{c|}{median-IGD}                                   & \multicolumn{3}{c|}{median-GD}                          \\ \cline{3-11}
                                  &                                      & MOEA/D             & MOEA/D-IR2-RF & \textbf{$p$}-value & MOEA/D    & MOEA/D-IR2-RF          & \textbf{$p$}-value  & MOEA/D    & MOEA/D-IR2-RF & \textbf{$p$}-value \\ \hline
ZDT1                              & 50                                                        & 0.534166          & \textbf{0.592069} & 2.04E-06         & 0.105212          & \textbf{0.085742} & 1.04E-02         & 0.117213          & \textbf{0.056263} & 1.40E-06         \\
ZDT2                              & 50                                                        & 0.162505          & \textbf{0.208247} & 2.91E-03         & 0.161240          & \textbf{0.109551} & 3.17E-02         & 0.163532          & \textbf{0.086702} & 2.43E-05         \\
ZDT3                              & 50                                                        & 0.900764          & \textbf{0.951958} & 3.96E-04         & 0.085657          & \textbf{0.055298} & 7.45E-03         & 0.065226          & \textbf{0.045721} & 1.64E-04         \\
ZDT4                              & 50                                                        & \textit{0.183939} & \textbf{0.254727} & 9.10E-01         & \textit{0.521270} & \textbf{0.405933} & 7.34E-01         & \textbf{0.636865} & \textit{0.786373} & 9.70E-01         \\
ZDT6                              & 50                                                        & \textit{0.000000} & \textbf{0.001219} & 2.00E-01         & \textit{0.668449} & \textbf{0.581419} & 1.32E-01         & 0.725946          & \textbf{0.659610} & 3.82E-02           \\\hline
\end{tabular}}
\end{table*}

\subsection{Other Multi- and Many-Objective Problems}

In this subsection, we analyze the performance of NSGA-III-IR2-RF and MOEA/D-IR2-RF on WFG and DTLZ problems ranging from 3 to 5 objectives. Earlier studies \cite{sukrit-ann-repair-gecco,sukrit-coin-ann} clearly indicate that the hypervolume improvement is significant in early generations but the difference reduces as the solutions approach the Pareto front. It is relevant to discuss the convergence in early generations as well, since the number of available function evaluations is limited in many real-world problems. Table~\ref{tab:nsga-iii-wfg} shows the analysis at generation $t=40$ with median hypervolume values $H^t$ and $H_R^t$ for NSGA-III and NSGA-III-IR2-RF, respectively. Another column, named \textit{Recovery}, depicts the generation ($t+\Delta t$) in which NSGA-III achieves a hypervolume equivalent to that of NSGA-III-IR2-RF at generation $t$. Then, the percentage savings in function evaluations ($S$) by the IR2 approach is calculated using Equation~\ref{eq:S-metric}. 
\begin{table*}[hbt]
\caption{Median Hypervolume metric and associated percentage savings in function evaluations ($S$-metric) on 3- and 4-objective WFG problems, solved using NSGA-III and NSGA-III-IR2-RF. Best performing framework is marked in bold.}
\label{tab:nsga-iii-wfg}
\centering
\begin{tabular}{|c|c|c|c|c|c|c|}
\hline
Problem-(M) &  {Generation ($t$)} &  {NSGA-III ($H^t$)} &  {NSGA-III-IR2-RF ($H_R^t$)} &  {\textbf{$p$}-value} &  {Recovery w/o repair ($t+\Delta t$)} &  {Saving ($S$\%)} \\ \hline
WFG1-3               & 40                                      & \textit{0.005261}                              & \textbf{0.005278}                        & 4.51E-01                             & 40                                                & 0.0                                      \\
WFG2-3               & 40                                      & \textit{0.781701}                     & \textbf{0.793153}                        & 2.91E-01                              & 47                                                 & 17.5                                     \\
WFG3-3               & 40                                      & 0.503857                              & \textbf{0.519115}                        & 1.73E-05                             & 55                                                 & 37.5                                     \\
WFG4-3               & 40                                      & 0.302729                              & \textbf{0.336154}                        & 1.41E-06                             & 75                                                 & 87.5                                     \\
WFG5-3               & 40                                      & \textbf{0.252149}                     & 0.248024                                 & 4.18E-02                             & 39                                                 & -2.5                                     \\
WFG6-3               & 40                                      & 0.225263                              & \textbf{0.273883}                        & 5.11E-06                             & 69                                                 & 72.5                                     \\
WFG7-3               & 40                                      & 0.275361                              & \textbf{0.309183}                        & 1.41E-06                             & 57                                                 & 42.5                                     \\
WFG8-3               & 40                                      & 0.224801                              & \textbf{0.240901}                        & 1.90E-04                             & 55                                                 & 37.5                                     \\
WFG9-3               & 40                                      & 0.284691                              & \textbf{0.327777}                        & 2.87E-05                             & 72                                                 & 80.0                                       \\\hline\hline
WFG1-4               & 40                                      & \textbf{0.057422}                              & \textit{0.052123}                        & 6.51E-01                             & 38                                                & -5.0                                      \\
WFG2-4               & 40                                      & 0.863800                              & \textbf{0.887670}                        & 7.57E-05                             & 52                                                 & 30.0                                       \\
WFG3-4               & 40                                      & 0.507060                              & \textbf{0.529724}                        & 1.41E-06                             & 65                                                 & 62.5                                     \\
WFG4-4               & 40                                      & 0.408224                              & \textbf{0.458472}                        & 1.41E-06                             & 79                                                 & 97.5                                     \\
WFG5-4               & 40                                      & \textbf{0.347052}                     & \textit{0.344402}                        & 9.73E-02                             & 40                                                 & 0.0                                        \\
WFG6-4               & 40                                      & 0.315560                              & \textbf{0.407118}                        & 1.41E-06                             & 93                                                 & 132.5                                    \\
WFG7-4               & 40                                      & 0.379668                              & \textbf{0.446049}                        & 1.41E-06                             & 68                                                 & 70.0                                       \\
WFG8-4               & 40                                      & 0.303973                              & \textbf{0.331764}                        & 1.41E-06                             & 60                                                 & 50.0                                       \\
WFG9-4               & 40                                      & 0.410231                              & \textbf{0.423810}                        & 1.04E-02                             & 48                                                 & 20.0           \\\hline                           
\end{tabular}
\end{table*}

The results in Table~\ref{tab:nsga-iii-wfg} indicate that the effectiveness of the proposed IR2 operator when used with the NSGA-III algorithm. The proposed algorithm performs worse only in problem WFG5 ($M=3$). This problem is extremely difficult and has deception, which makes it fundamentally difficult for any kind of learning-based repair to yield better performance. However, it is also evident that the degradation in performance on that instance is much lower than the average improvements in most other test instances. 
A similar study is done for MOEA/D for three- and four-objective problems to demonstrate the effectiveness of the proposed IR2 operator when integrated with MOEA/D. The results are tabulated in Table~\ref{tab:moead-wfg}, where the trend is in line with the results on NSGA-III mentioned in Table~\ref{tab:nsga-iii-wfg}. The only difficulty MOEA/D-IR2-RF faces is on solving WFG5. However, the difference is not statistically significant, as indicated by the $p$-value. 
\begin{table*}[hbt]
\caption{Median Hypervolume metric and associated percentage savings in function evaluations ($S$-metric) on 3- and 4-objective WFG problems, solved using MOEA/D and MOEA/D-IR2-RF. Best performing framework is marked in bold.}
\label{tab:moead-wfg}
\centering
\begin{tabular}{|c|c|c|c|c|c|c|}
\hline
Problem-($M$) &  {Generation ($t$)} &  {MOEA/D ($H^t$)} &  {MOEA/D-IR2-RF ($H_R^t$)} &  {\textbf{$p$}-value} &  {Recovery w/o repair ($t+\Delta t$)} &  {Saving ($S$\%)} \\ \hline
WFG1-3               & 40                                      & 0.022009                            & \textbf{0.170737}                      & 1.23E-05                             & 83                                               & 107.5                                    \\
WFG2-3               & 40                                      & \textit{0.576799}                   & \textbf{0.582327}                      & 5.97E-01                              & 46                                               & 15.0                                       \\
WFG3-3               & 40                                      & \textit{0.430079}                   & \textbf{0.439573}                      & 3.46E-01                             & 47                                               & 17.5                                     \\
WFG4-3               & 40                                      & 0.203112                            & \textbf{0.231028}                      & 6.11E-06                             & 78                                               & 95.0                                       \\
WFG5-3               & 40                                      & \textbf{0.141332}                   & \textit{0.132609}                      & 3.27E-01                             & 32                                               & -20.0                                      \\
WFG6-3               & 40                                      & 0.110332                            & \textbf{0.138521}                      & 6.65E-03                             & 65                                               & 62.5                                     \\
WFG7-3               & 40                                      & 0.150090                            & \textbf{0.192004}                      & 4.26E-06                             & 134                                              & 235.0                                      \\
WFG8-3               & 40                                      & 0.128467                            & \textbf{0.145331}                      & 1.95E-02                             & 59                                               & 47.5                                     \\
WFG9-3               & 40                                      & 0.134948                            & \textbf{0.188236}                      & 1.60E-04                             & 99                                               & 147.5                                    \\\hline\hline
WFG1-4               & 40                                      & 0.069468                            & \textbf{0.196936}                      & 1.41E-06                             & 156                                              & 290.0                                      \\
WFG2-4               & 40                                      & \textit{0.591580}                   & \textbf{0.607856}                      & 3.46E-01                              & 53                                               & 32.5                                     \\
WFG3-4               & 40                                      & 0.424510                            & \textbf{0.452893}                      & 6.90E-04                             & 56                                               & 40.0                                       \\
WFG4-4               & 40                                      & 0.271415                            & \textbf{0.300305}                      & 2.20E-04                             & 68                                               & 70.0                                       \\
WFG5-4               & 40                                      & \textbf{0.194265}                   & \textit{0.183675}                      & 5.22E-01                             & 36                                               & -10.0                                      \\
WFG6-4               & 40                                      & 0.156274                            & \textbf{0.196048}                      & 7.90E-04                             & 61                                               & 52.5                                     \\
WFG7-4               & 40                                      & 0.206851                            & \textbf{0.249464}                      & 2.04E-06                             & 87                                               & 117.5                                    \\
WFG8-4               & 40                                      & 0.162210                            & \textbf{0.184629}                      & 5.20E-04                             & 63                                               & 57.5                                     \\
WFG9-4               & 40                                      & 0.177231                            & \textbf{0.248910}                      & 2.05E-05                             & 96                                               & 140.0            \\\hline                         
\end{tabular}
\end{table*}

Table~\ref{tab:nsga-iii-dtlz} shows the median hypervolume results obtained from NSGA-III and NSGA-III-IR2-RF on three- to five-objective problems. The results are recorded at an intermediate generation and the final generation when the optimizer is terminated. Here, the intermediate generation is chosen as $t+100$, where $t$ is the generation at which the first non-zero hypervolume is achieved. The table clearly shows that the performance of NSGA-III-IR2-RF is better at the intermediate generation, however the performance is either better, or, at par in the final generation. These results support two important arguments. 
\begin{enumerate}
    \item For a problem in which PO solutions can be achieved by NSGA-III (the base EMO/EMaO algorithm), NSGA-III will eventually catch up with NSGA-III-IR2-RF in terms of any performance metric measured. However, there are problems like WFG6, WFG9, etc. where the population may stagnate before achieving the PO solution set. In these problems, NSGA-III-IR2 achieve equivalent or better results than NSGA-III.
    \item For real-world problems where the available function evaluations are limited, it is important to build an optimizer that can produce a better solution set (more converged) quickly (or, at any intermediate generation). The performance at the intermediate generations, as shown on ZDT, WFG and DTLZ problems, reveals that the proposed learning based-IR2 operator achieves better convergence in most test instances, thus demonstrating its efficacy. 
\end{enumerate}
\begin{table*}[hbt]
\caption{Hypervolume values for DTLZ1--4 ($M$=3,4,5), comparing NSGA-III without repair and with the proposed IR2 repair operator. The intermediate generation is chosen as 100 generations after the first non-zero hypervolume is achieved.
Best performing framework is marked in bold and statistically similar results are marked in italics.}
\label{tab:nsga-iii-dtlz}
\centering
\begin{tabular}{|c|c|c|c|c|c|c|c|c|}
\hline
\multicolumn{1}{|c|}{\multirow{2}{*}{Problem-($M$)}} & \multicolumn{4}{c|}{Intermediate Generation   HV}              & \multicolumn{4}{c|}{End Generation HV}            \\ \cline{2-9}
\multicolumn{1}{|c|}{}                             & Generation & NSGA-III          & NSGA-III-IR2-RF       & p-value  & Generation & NSGA-III          & NSGA-III-IR2-RF       & p-value  \\\hline
DTLZ1-3                                          & 550        & \textit{0.067417} & \textbf{0.315313} & 3.27E-01 & 1000 &  \textit{1.001629} & \textbf{1.002444} & 1.52E-01 \\
DTLZ2-3                                          & 103        & 0.433458          & \textbf{0.438385} & 1.41E-06 & 1000 & \textbf{0.447479} & \textit{0.447471} & 8.51E-01 \\
DTLZ3-3                                          & 687        & \textit{0.203485} & \textbf{0.361433} & 5.72E-01 & 1000 & \textit{0.434177} & \textbf{0.439590} & 3.86E-01 \\
DTLZ4-3                                          & 109        & 0.435713          & \textbf{0.439147} & 3.48E-02 & 1000 & \textbf{0.447480} & \textit{0.447469} & 1.87E-01 \\\hline\hline
DTLZ1-4                                          & 443        & \textit{0.032100} & \textbf{0.299945} & 1.18E-01 & 1000 & \textit{1.009436} & \textbf{1.009460} & 4.74E-01 \\
DTLZ2-4                                          & 100        & 0.589208          & \textbf{0.601760} & 1.41E-06 & 1000 & \textbf{0.618473} & \textit{0.618437} & 7.34E-01 \\
DTLZ3-4                                          & 643        & 0.195940          & \textbf{0.534411} & 5.94E-03 & 1000 & 0.606276          & \textbf{0.612803} & 2.27E-03 \\
DTLZ4-4                                          & 107        & 0.595864          & \textbf{0.609001} & 1.41E-06 & 1000 & \textit{0.618445} & \textbf{0.618466} & 9.40E-01 \\\hline\hline
DTLZ1-5                                          & 386        & \textit{0.657787} & \textbf{0.830186} & 4.29E-01 & 1000 & \textbf{1.009373} & \textit{1.009368} & 7.92E-01 \\
DTLZ2-5                                          & 100        & 0.702793          & \textbf{0.721317} & 1.41E-06 & 1000 & \textit{0.743545} & \textbf{0.743554} & 9.10E-01 \\
DTLZ3-5                                          & 585        & \textit{0.526269} & \textbf{0.567091} & 3.86E-01 & 1000 & \textit{0.736317} & \textbf{0.737042} & 8.80E-01 \\
DTLZ4-5                                          & 105        & 0.709817          & \textbf{0.733610} & 1.41E-06 & 1000 & \textit{0.743692} & \textbf{0.743728} & 1.42E-01 \\\hline
\end{tabular}
\end{table*}


\section{Discussion on IR2 Operator} \label{sec:discussion}

Based on the results of multiple two- to five-objective problems, it is evident that algorithms employing the IR2 operator perform much better than the original algorithms in most cases; however, there are some instances where the IR operator does not change the original algorithm's performance. There are certain attributes of problems that can make it harder for an EMO/EMaO without learning-based repair to perform well. In the following subsections, we discuss some of these characteristics, where a modified IR2 based algorithms yield considerable improvement, like multi-modality, bias and variable linkages. Further, we discuss the effect of boundary points in many-objective problems, and the effect and importance of early termination of the learning algorithm. 

\subsection{Multi-Modality of Pareto-optimal Set}

\begin{figure}[hbt]
    \centering
\begin{subfigure}[b]{0.49\linewidth}
    \centering
    \includegraphics[width=1.05\linewidth]{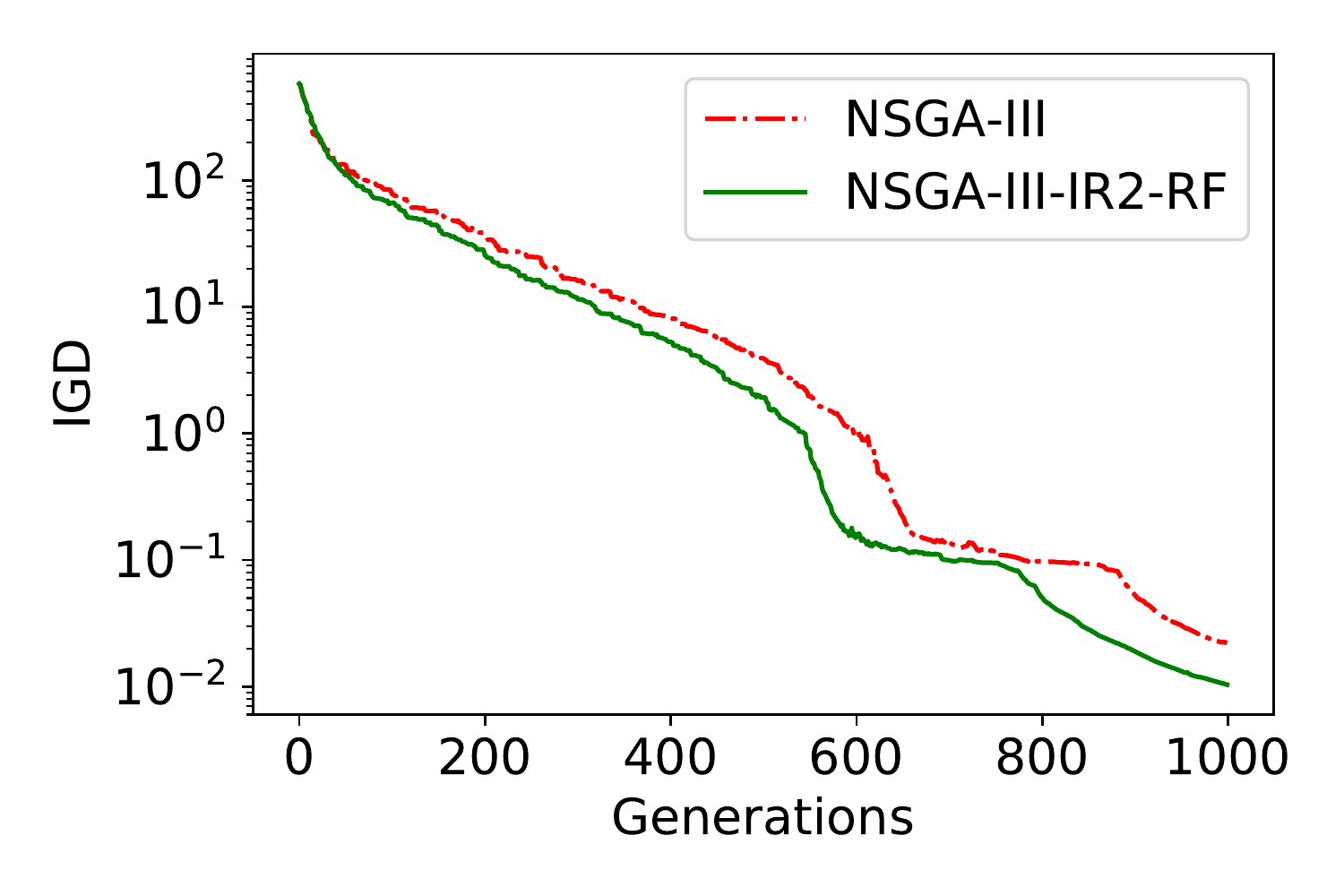}
    \caption{DTLZ3 ($M=4$)}
    \label{fig:IGD-DTLZ3-4}
    \end{subfigure}
\begin{subfigure}[b]{0.49\linewidth}
    \centering
    \includegraphics[width=1.05\linewidth]{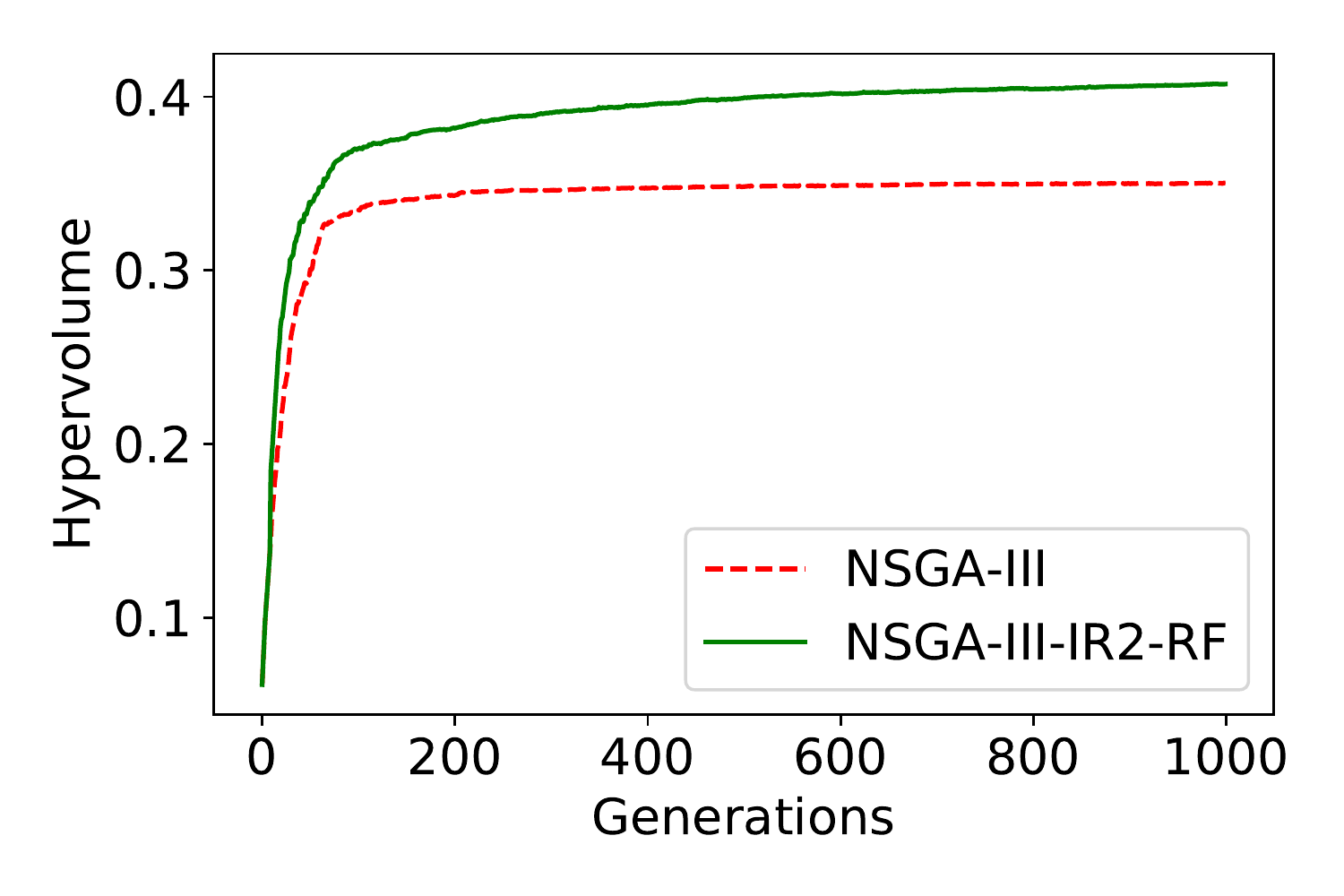}
    \caption{WFG9 ($M=3$)}
    \label{fig:HV-WFG9}
    \end{subfigure}
\caption{Median IGD plot for DTLZ3 and median HV plot for WFG9 (Multi-modal problems).}
\end{figure}

Some test problem problems exhibiting multi-modality are DTLZ3, WFG4 and WFG9 \cite{WFG}. The generation-wise median IGD plot of DTLZ3 and median hypervolume plot of WFG9 are shown in Figures~\ref{fig:IGD-DTLZ3-4} and \ref{fig:HV-WFG9}, respectively. For DTLZ3, IGD is chosen since the median hypervolume remains zero until almost 600 generations in a 1,000 generation run. In Figure~\ref{fig:IGD-DTLZ3-4}, it is evident that the performance of NSGA-III-IR2-RF is better than that of NSGA-III at all times from start to end. Similarly, Figure~\ref{fig:HV-WFG9} shows that the use of the IR2 operator helps NSGA-III converge to a better solution for problem WFG9. It is clear that without the IR2 operator, NSGA-III converges to a smaller hypervolume value. 

\begin{figure}[hbt]
\begin{subfigure}[b]{0.49\linewidth}
    \centering    
    \includegraphics[width=1.05\linewidth]{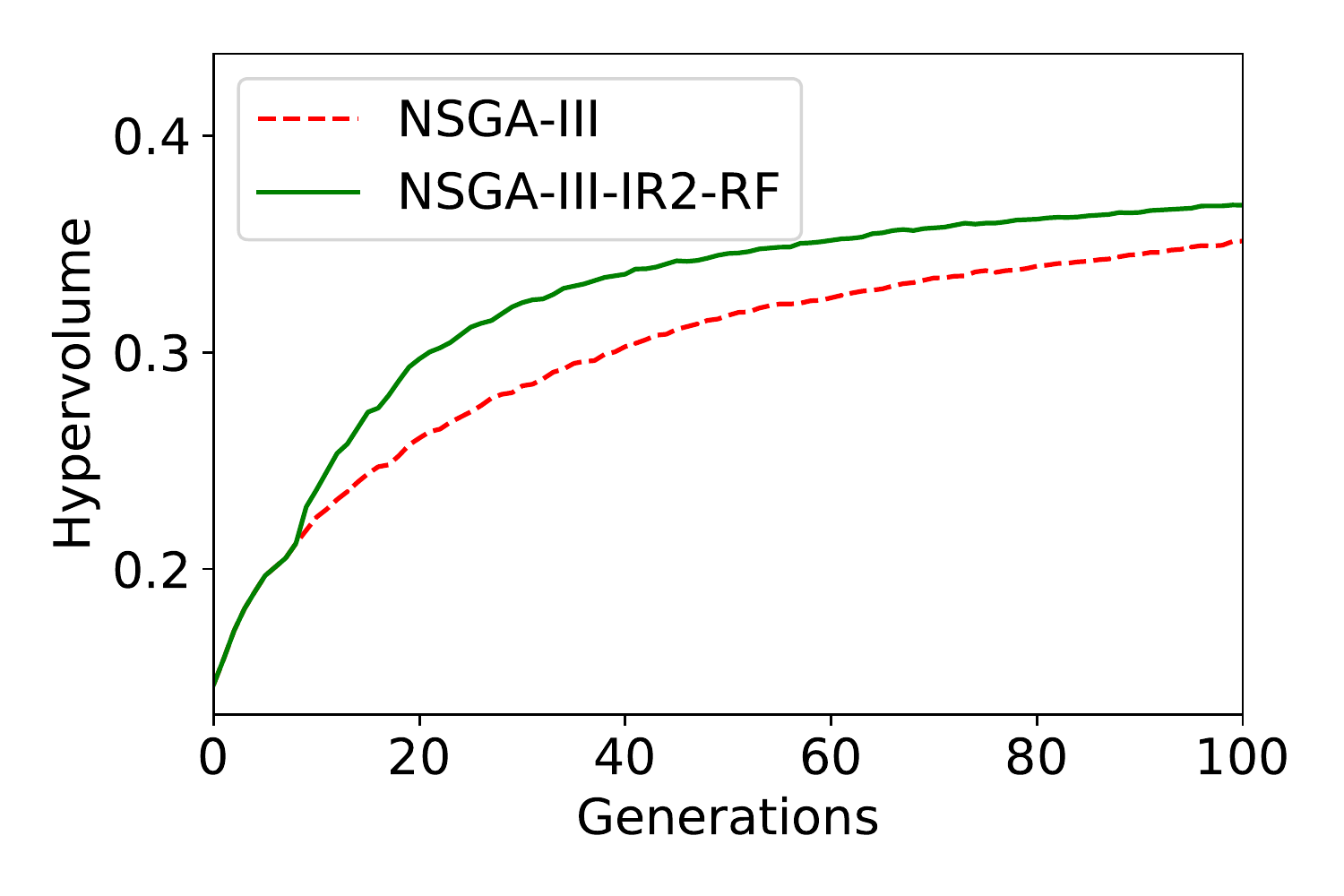}
    \caption{WFG4 ($M=3$)}
    \label{fig:WFG4-3-normal}
    \end{subfigure}
\begin{subfigure}[b]{0.49\linewidth}
    \centering
    \includegraphics[width=1.05\linewidth]{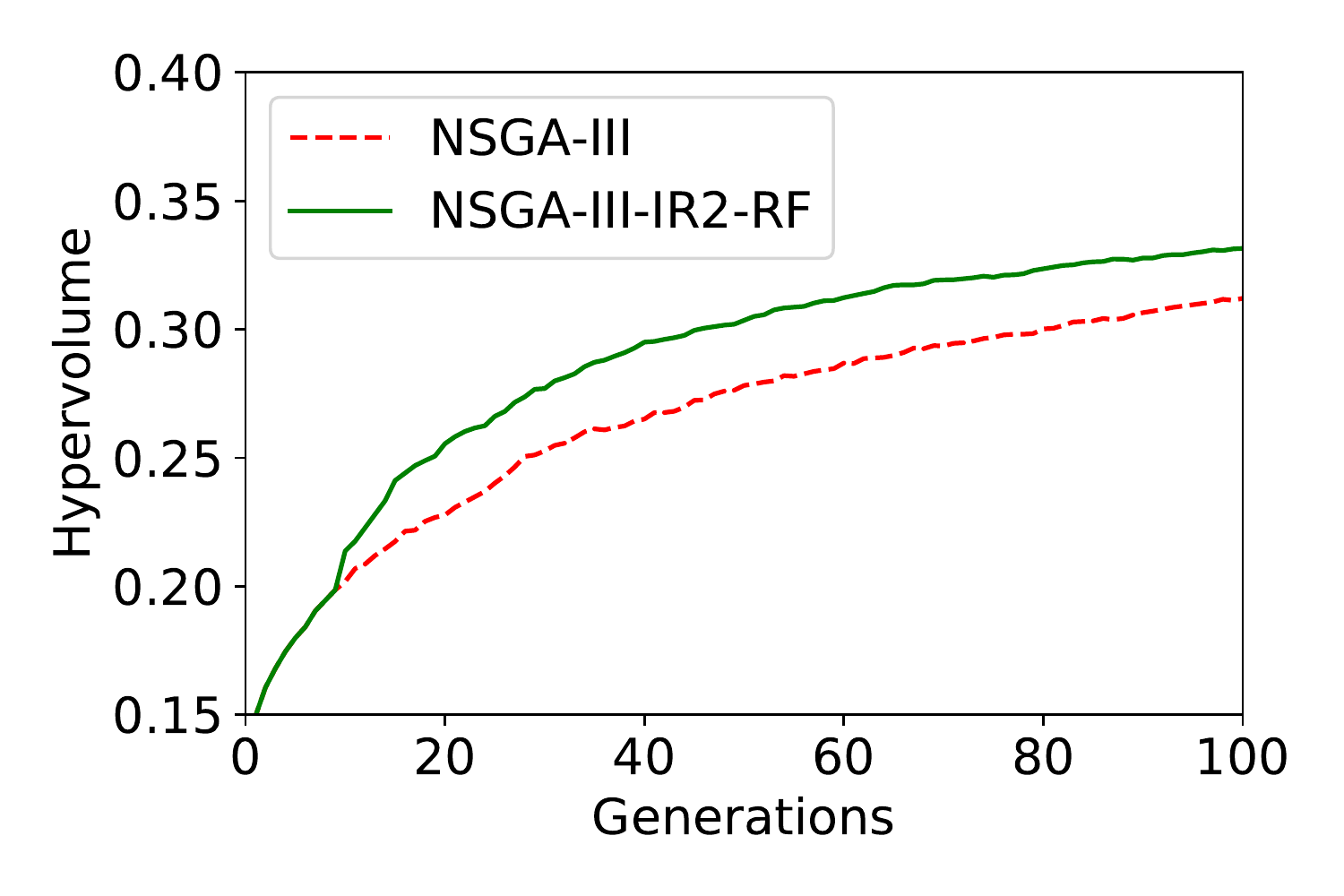}
    \caption{Modified WFG4 ($M=3$)}
    \label{fig:WFG4-3-special}
\end{subfigure}
\caption{Median HV plots for WFG4 (with standard and difficult versions).}
\label{fig:WFG4}
\end{figure}

Motivated from results on these and other problems, we further increase the difficulty of WFG4 problem. The multi-modality of the problem is dependent on three parameters: $A$, $B$ and $C$, whose standard values are 30.0, 10.0 and 0.35, respectively. Earlier study \cite{WFG} suggested that higher values of $A$ and smaller values of $B$ will create more difficult problems. Hence, we define the new problem as modified-WFG4 with parameter values as 70.0, 5.0 and 0.35. The HV plots for both WFG4 and modified-WFG4 are shown in Figure~\ref{fig:WFG4}.
Both hypervolume values in Figure~\ref{fig:WFG4-3-special} are comparatively lower than their corresponding values in Figure~\ref{fig:WFG4-3-normal}, but NSGA-III-IR2-RF is able to yield a performance improvement even in the more difficult version of WFG4. 

\subsection{Bias in Pareto-optimal Set}

\begin{figure}[hbt]
    \centering
\begin{subfigure}[b]{0.49\linewidth}
    \centering
    \includegraphics[width=1.05\linewidth]{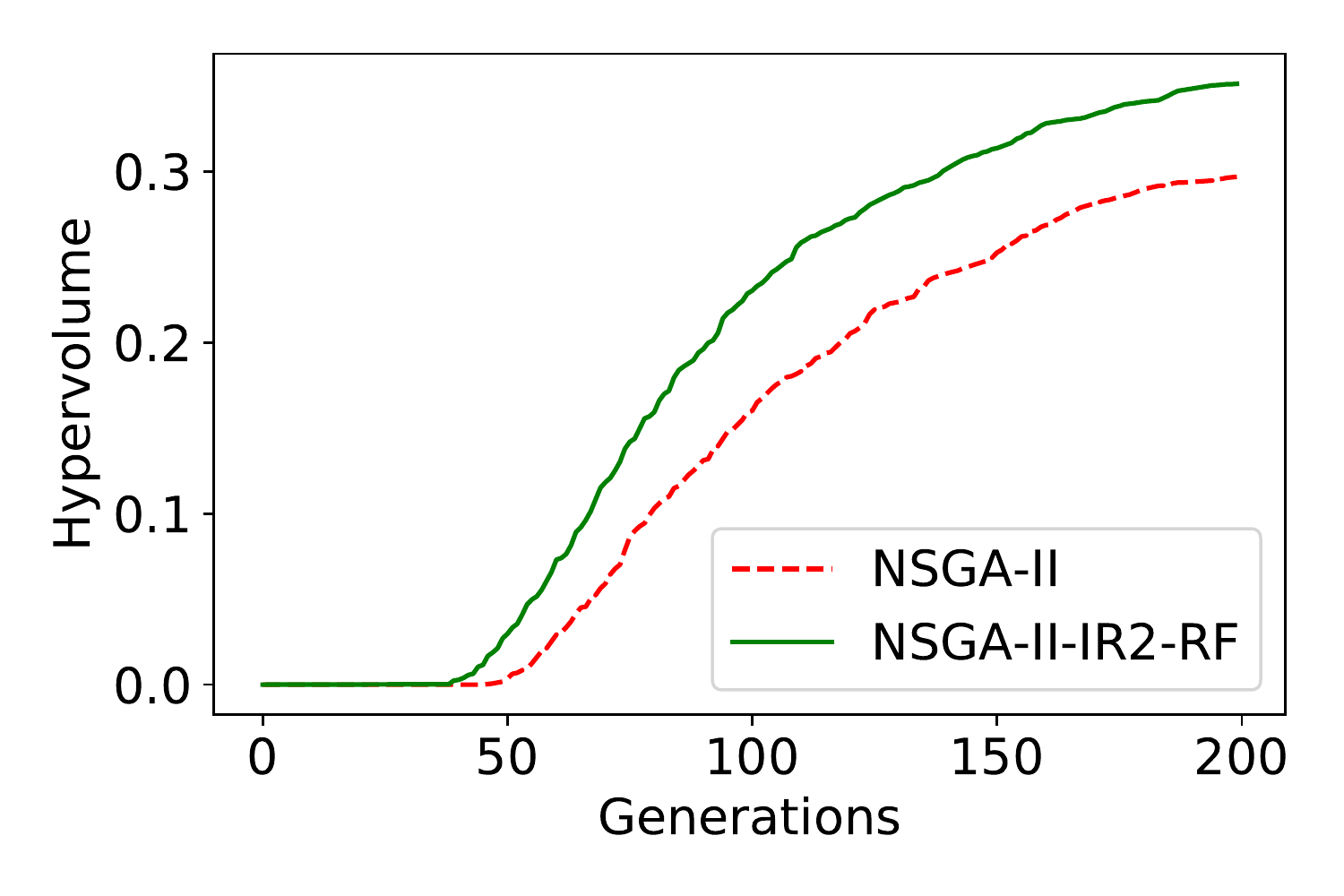}
    \caption{ZDT6}
    \label{fig:HV-ZDT6}
    \end{subfigure}
\begin{subfigure}[b]{0.49\linewidth}
    \centering
    \includegraphics[width=1.05\linewidth]{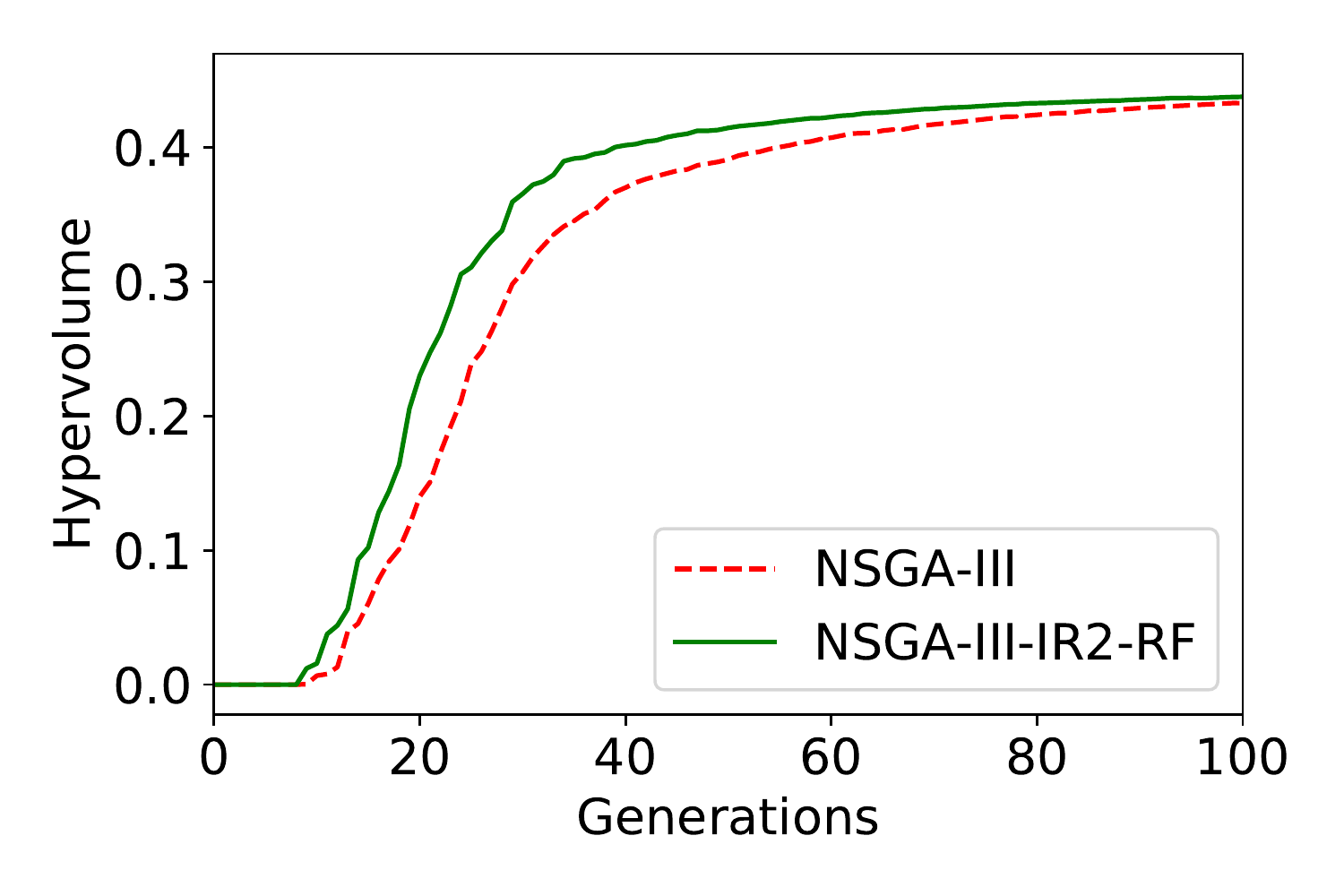}
    \caption{DTLZ4 ($M=3$)}
    \label{fig:HV-DTLZ4}
    \end{subfigure}
\caption{Median HV plots for biased problems: ZDT6 and DTLZ4.}
\end{figure}

Further, some known problems with bias in the objective space are ZDT6, DTLZ4, and WFG7 \cite{WFG}. The generation-wise median hypervolume plots for problems ZDT6 and DTLZ4 are shown in Figures~\ref{fig:HV-ZDT6} and \ref{fig:HV-DTLZ4}, respectively. The usefulness of the proposed IR2 operator is evident from these figures. Theoretically, it makes sense to have the advantage of a learning-based operator in biased problems, since the learning will help reach the optimal distribution of solutions faster once the right extreme points are identified. 

\begin{figure}[hbt]    
\begin{subfigure}[b]{0.49\linewidth}
    \centering    
    \includegraphics[width=1.05\linewidth]{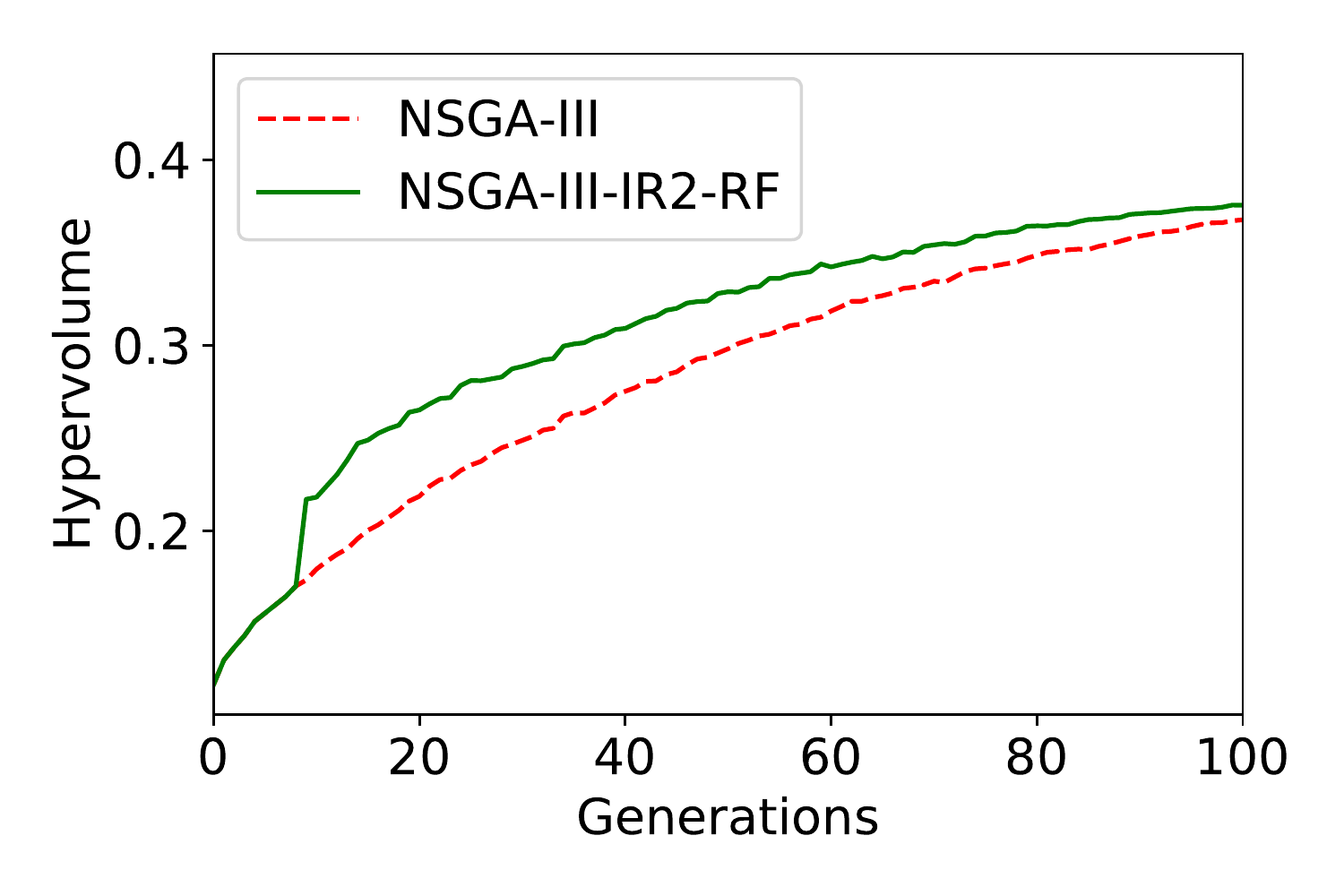}
    \caption{WFG7 ($M=3$)}
    \label{fig:WFG7-3-normal}
    \end{subfigure}
\begin{subfigure}[b]{0.49\linewidth}
    \centering
    \includegraphics[width=1.05\linewidth]{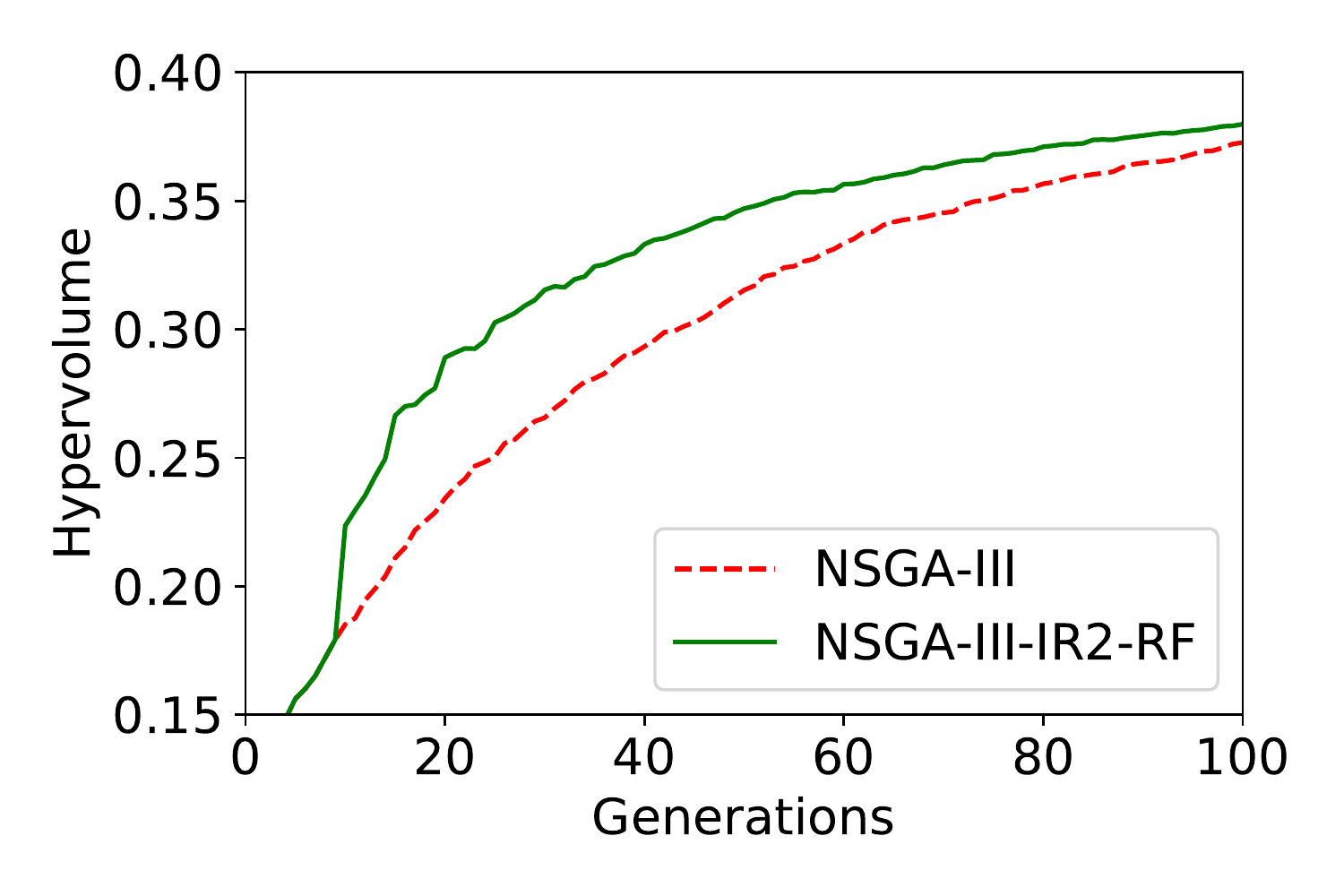}
    \caption{Modified WFG7 ($M=3$)}
    \label{fig:WFG7-3-special}
\end{subfigure}
\caption{Median HV plots for WFG7 (with standard and difficult versions).}
\label{fig:WFG7}
\end{figure}

As done for the multi-modal problem WFG4, the parameters of WFG7 problem are also modified since it has a parameter-dependent bias. The parameter $C$ in the WFG7 problem, which is suggested to be 50 in \cite{WFG}, is now increased to 100. We call this problem with $C=100$ the modified-WFG7. The HV plots of both versions of WFG7 are shown in Figure~\ref{fig:WFG7}. 

\subsection{Linked Variable Values in Pareto-optimal Set}

\begin{figure}[hbt]    
\begin{subfigure}[b]{0.49\linewidth}
    \centering    
    \includegraphics[width=1.05\linewidth]{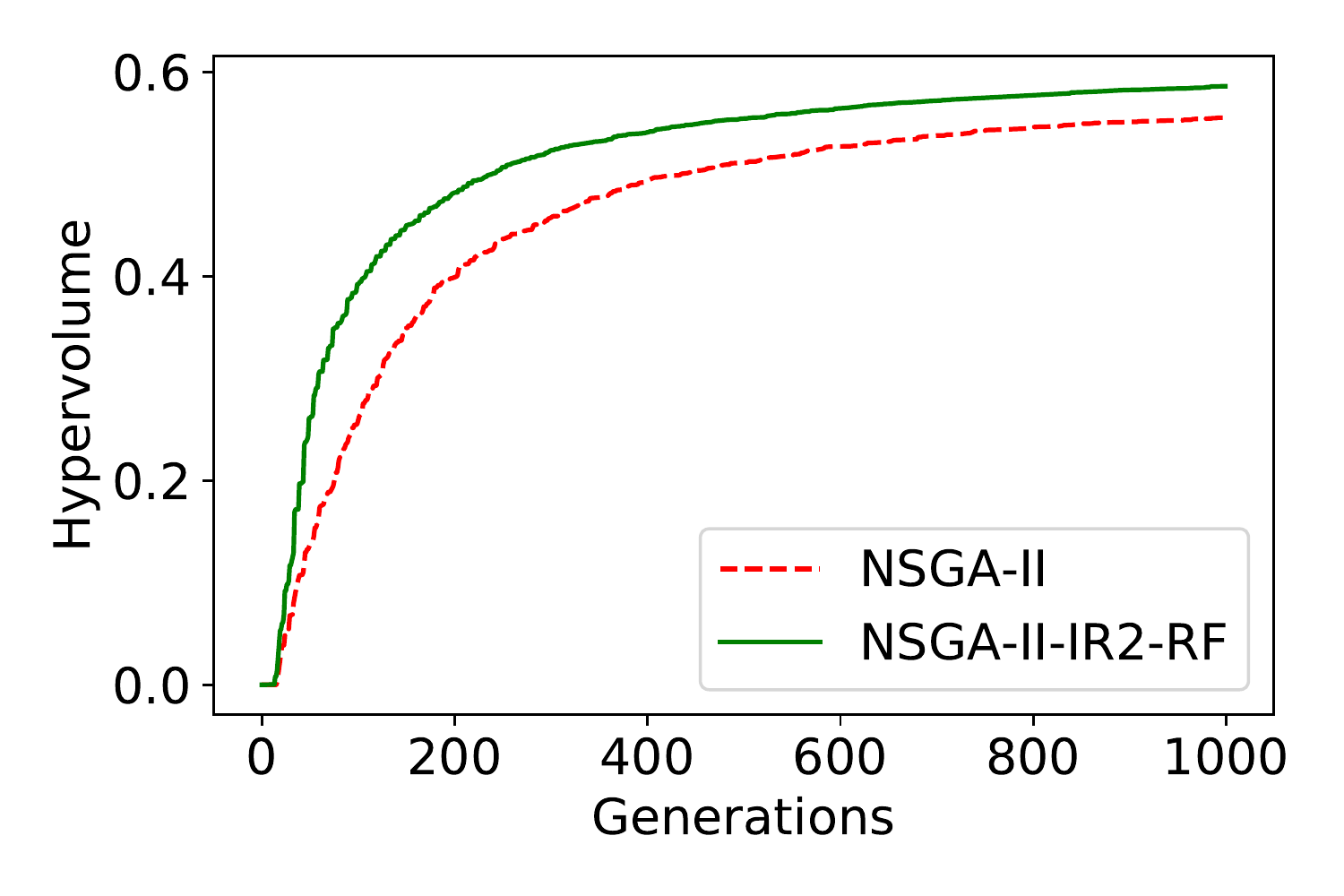}
    \caption{L1}
    \label{fig:HV-L1}
    \end{subfigure}
\begin{subfigure}[b]{0.49\linewidth}
    \centering
    \includegraphics[width=1.05\linewidth]{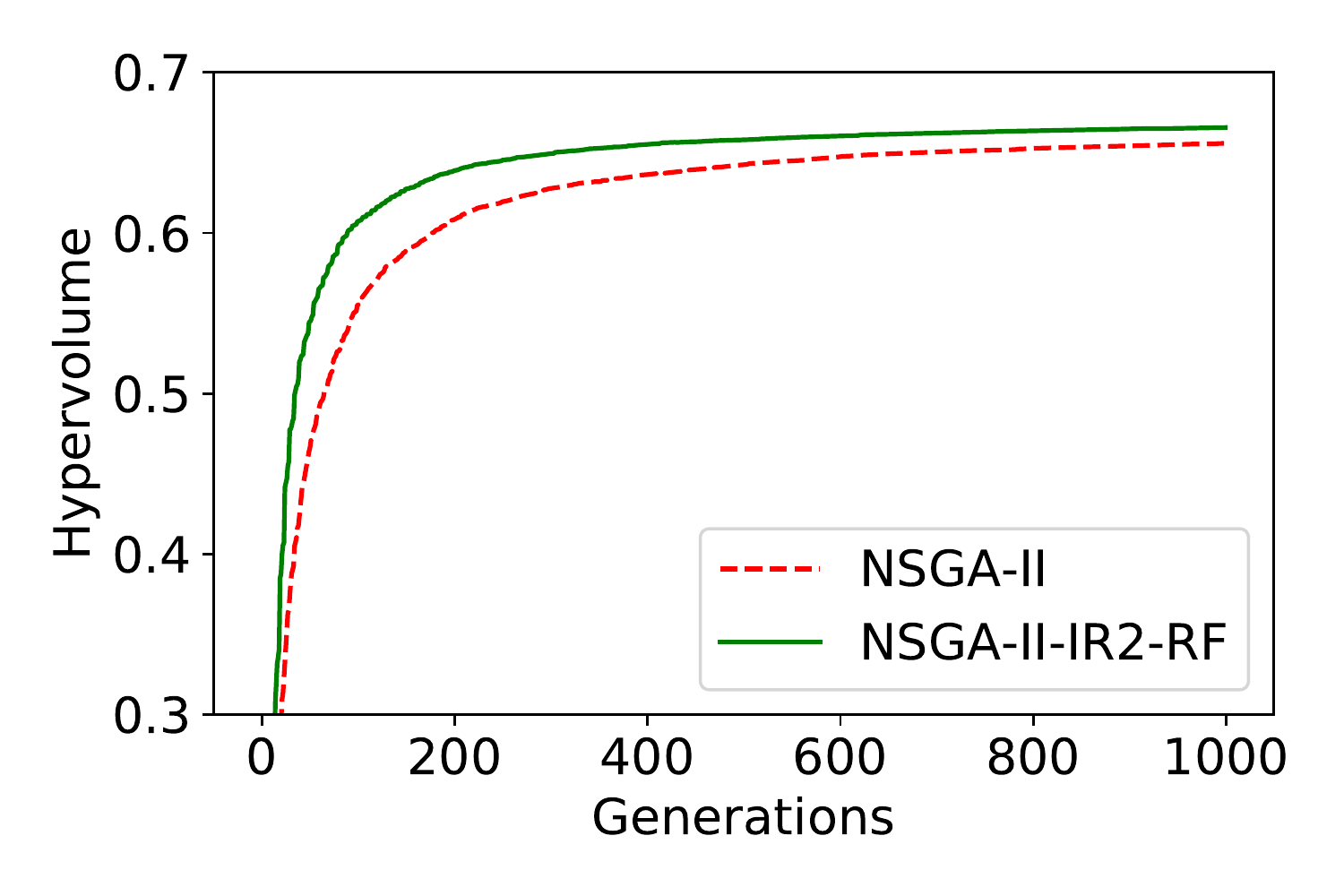}
    \caption{L2}
    \label{fig:HV-L2}
\end{subfigure}
\caption{Median HV plots for L1 and L2 problems with NSGA-II and NSGA-II-IR2-RF.}
\label{fig:HV-L}
\end{figure}

It is known that the common genetic operators, like SBX crossover and polynomial mutation operators, do not account for any inter-variable relationships or linkages. A crossover operator that can preserve variable linkages is the operator from  \textit{differential evolution} (DE). However, in problems with linked as well as in problems with independent variables, DE is not as efficient in gaining convergence \cite{sukrit-coin-ann}. In both of these cases, the SBX crossover operator with a learning-based IR operator proves to be effective. Two problems L1 and L2 \cite{sukrit-coin-ann} with linked variables are solved using NSGA-II and NSGA-II-IR2-RF. The generation-wise hypervolume plots are presented in Figures~\ref{fig:HV-L1} and \ref{fig:HV-L2}. These plots clearly depict that the IR operator proves effective in preserving the variable linkages. 

\subsection{Energy Method for Generating Reference Points}

Using the Das-Dennis method to generate reference points (parameters $M$ and $p$ taken from Table~\ref{tab:population-size}), 
the percentage of boundary points to the all reference points is tabulated in Table~\ref{tab:boundary-points}. It is evident from the table that the percentage of boundary points escalates quickly with the number of objectives. This can potentially affect the quality of learning as well. 
\begin{table}[hbt]
    \caption{Percentage boundary Das-Dennis points with $M$ objectives.}
    \label{tab:boundary-points}
    \centering
    \begin{tabular}{|c|c|c|c|c|} \hline
        $M$ & 2 & 3 & 4 & 5\\ \hline
        Boundary Points & 2\% & 37\% & 70\% & 93\% \\ \hline
    \end{tabular}
\end{table}
In order to explore this effect, the recently-proposed energy method \cite{Energy-TEC-2020} is used here to generate reference points and the benefit of learning is compared. Using the energy method in a layer-wise manner, we can control the proportion of boundary points. With $M=4$, 5 layers of reference points are generated with gaps in each layer as [16, 13, 11, 8, 2] (from outer to inner layer). By this method, 290 points are generated with only 32\% boundary points. This reduction from 70\% to 32\% boundary points (shifting from Das-Dennis to Energy Method) in turn reflects that the proportion of interior points has increased. 
\begin{figure}[hbt]    
\begin{subfigure}[b]{0.49\linewidth}
    \centering    
    \includegraphics[width=1.05\linewidth]{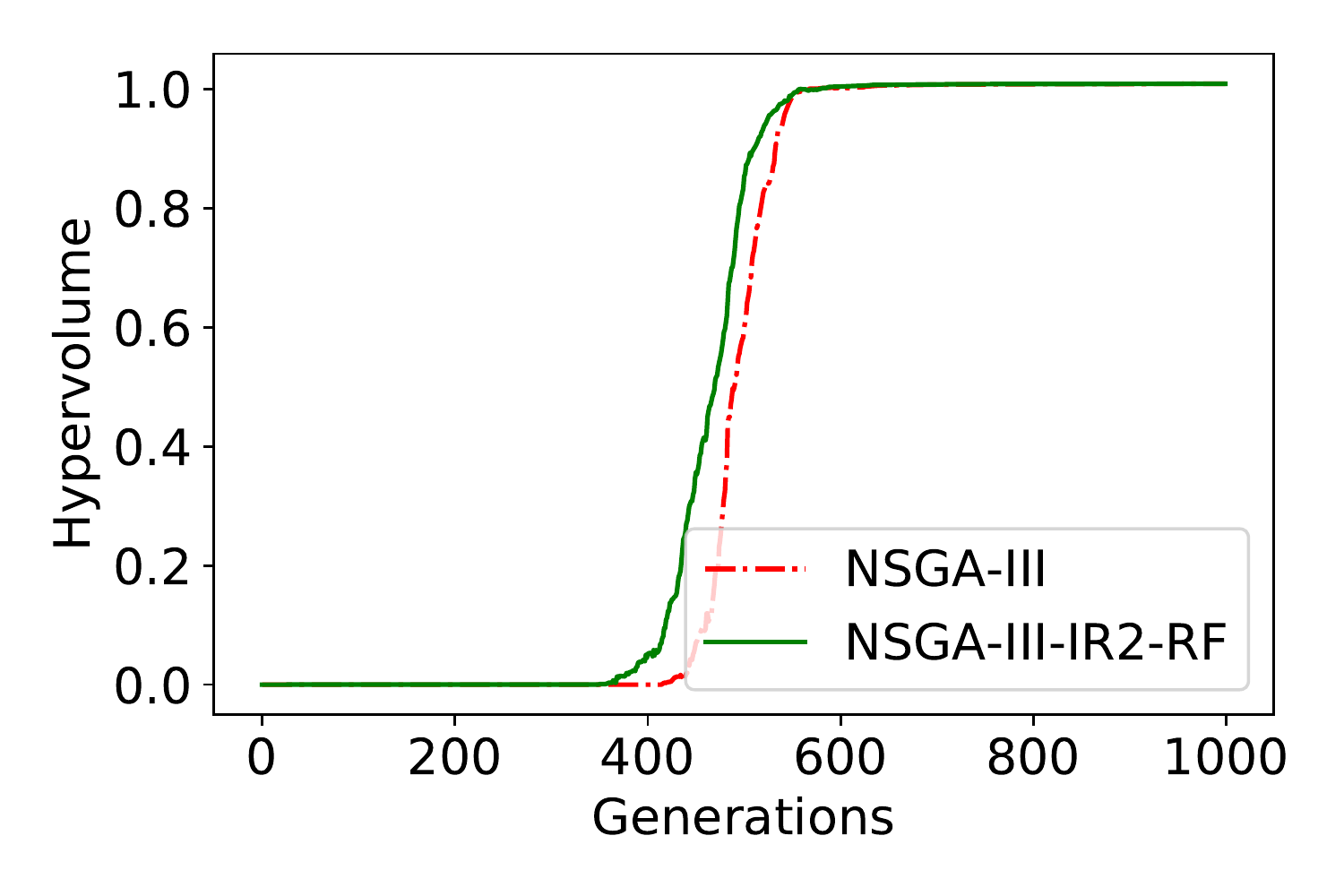}
    \caption{Das-Dennis method.}
    \label{fig:HV-Das-Dennis}
    \end{subfigure}
\begin{subfigure}[b]{0.49\linewidth}
    \centering
    \includegraphics[width=1.05\linewidth]{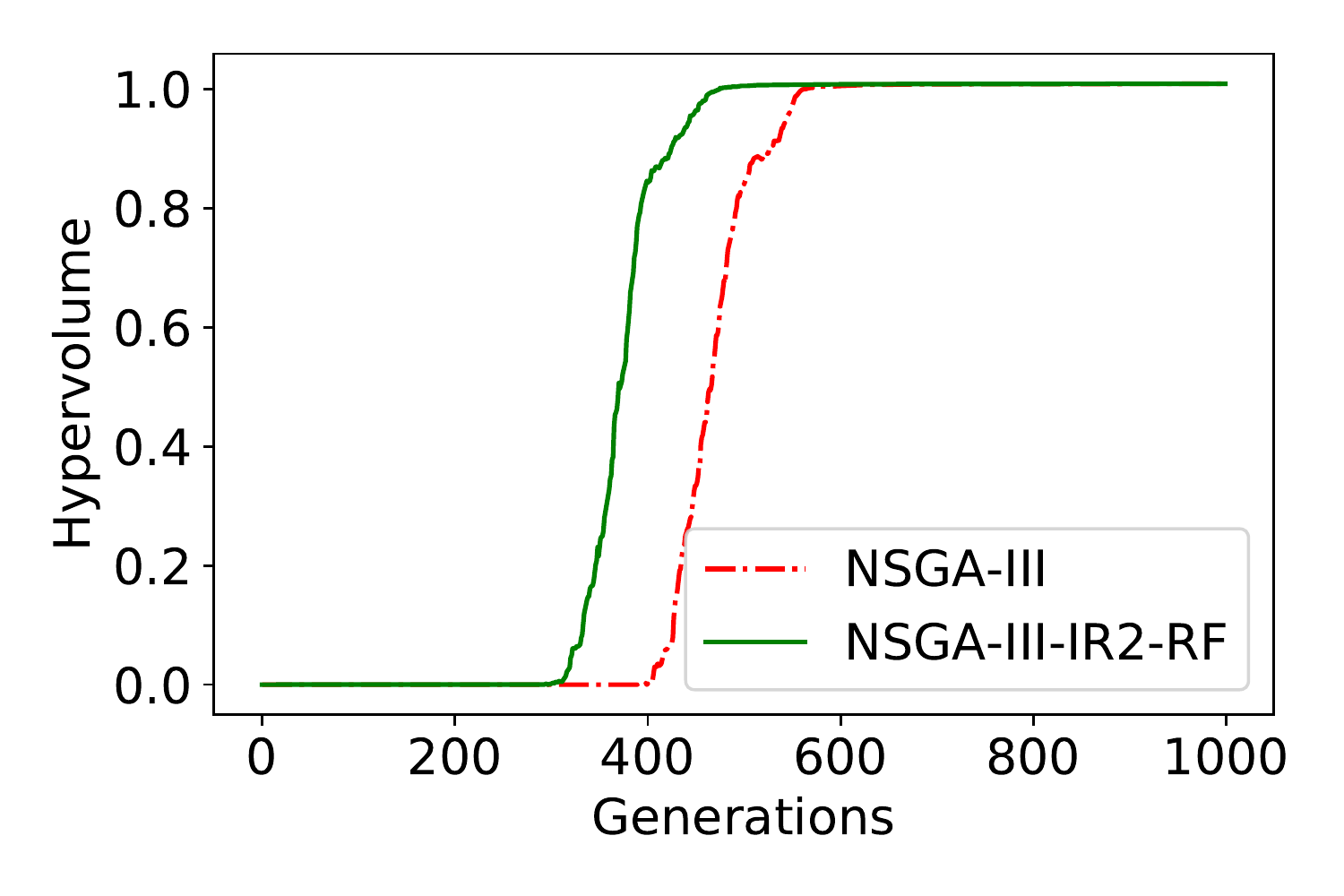}
    \caption{Layer-wise energy method.}
    \label{fig:HV-Energy}
\end{subfigure}
\caption{Median HV plots for DTLZ1 ($M=4$) problem solved by NSGA-III and NSGA-III-IR2-RF using reference points generated by Das-Dennis method and layer-wise energy Method.}
\label{fig:HV-DTLZ1}
\end{figure}

It can be seen in Figure~\ref{fig:HV-DTLZ1} that the performance difference is much larger in the case of the energy method compared to the Das-Dennis method, even though the population size is quite similar in both cases (286 and 290 in Figures~\ref{fig:HV-Das-Dennis} and \ref{fig:HV-Energy}, respectively). This motivates us to explore the effect of boundary points and interior points on quality of learning in future studies. 

\subsection{Limited Use of Learning} \label{sec:termination-discussion}

\begin{figure}[hbt]    
\begin{subfigure}[b]{0.49\linewidth}
    \centering    
    \includegraphics[width=1.05\linewidth]{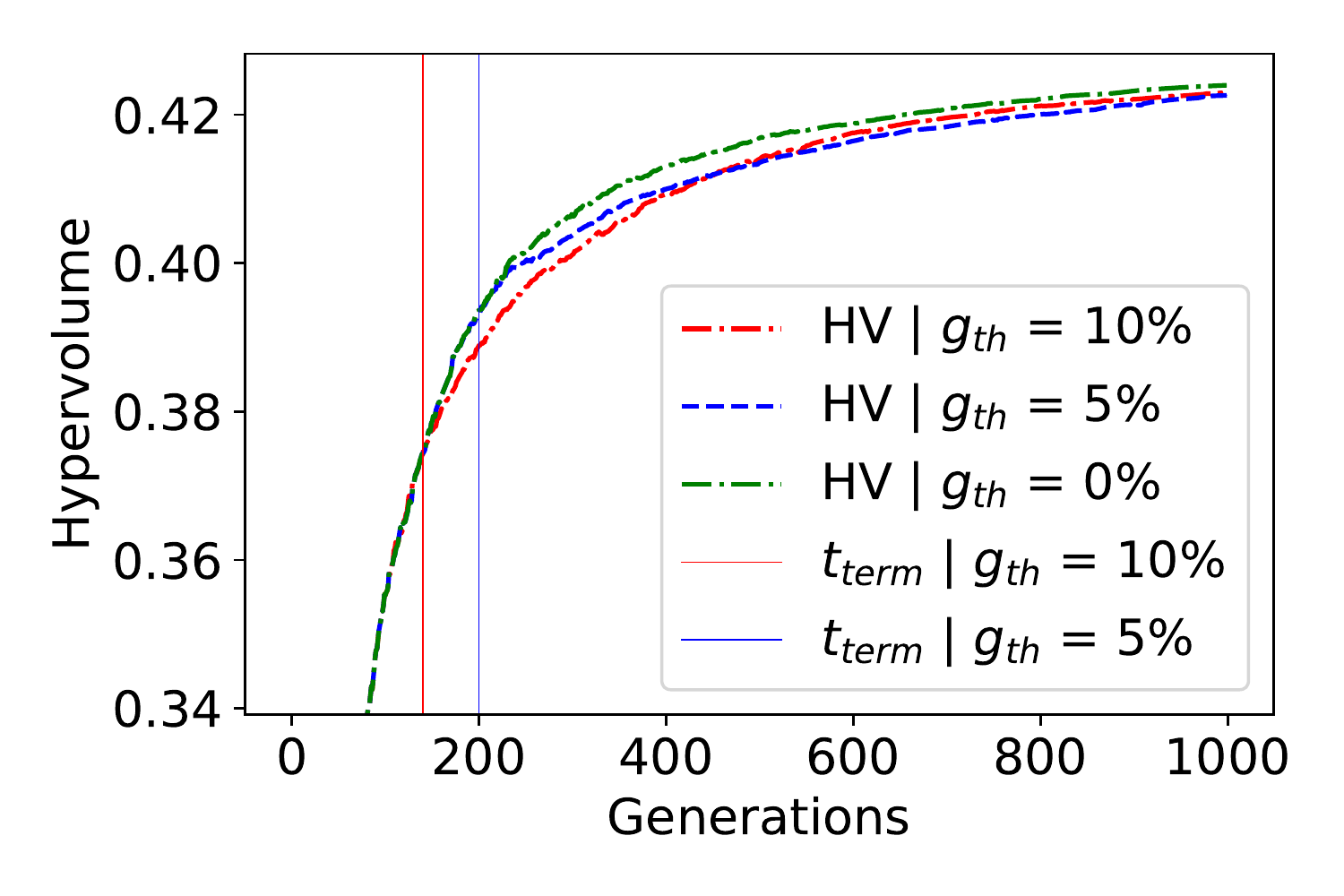}
    \caption{WFG6 ($M=3$)}
    \label{fig:HV-WFG6-term}
    \end{subfigure}
\begin{subfigure}[b]{0.49\linewidth}
    \centering
    \includegraphics[width=1.05\linewidth]{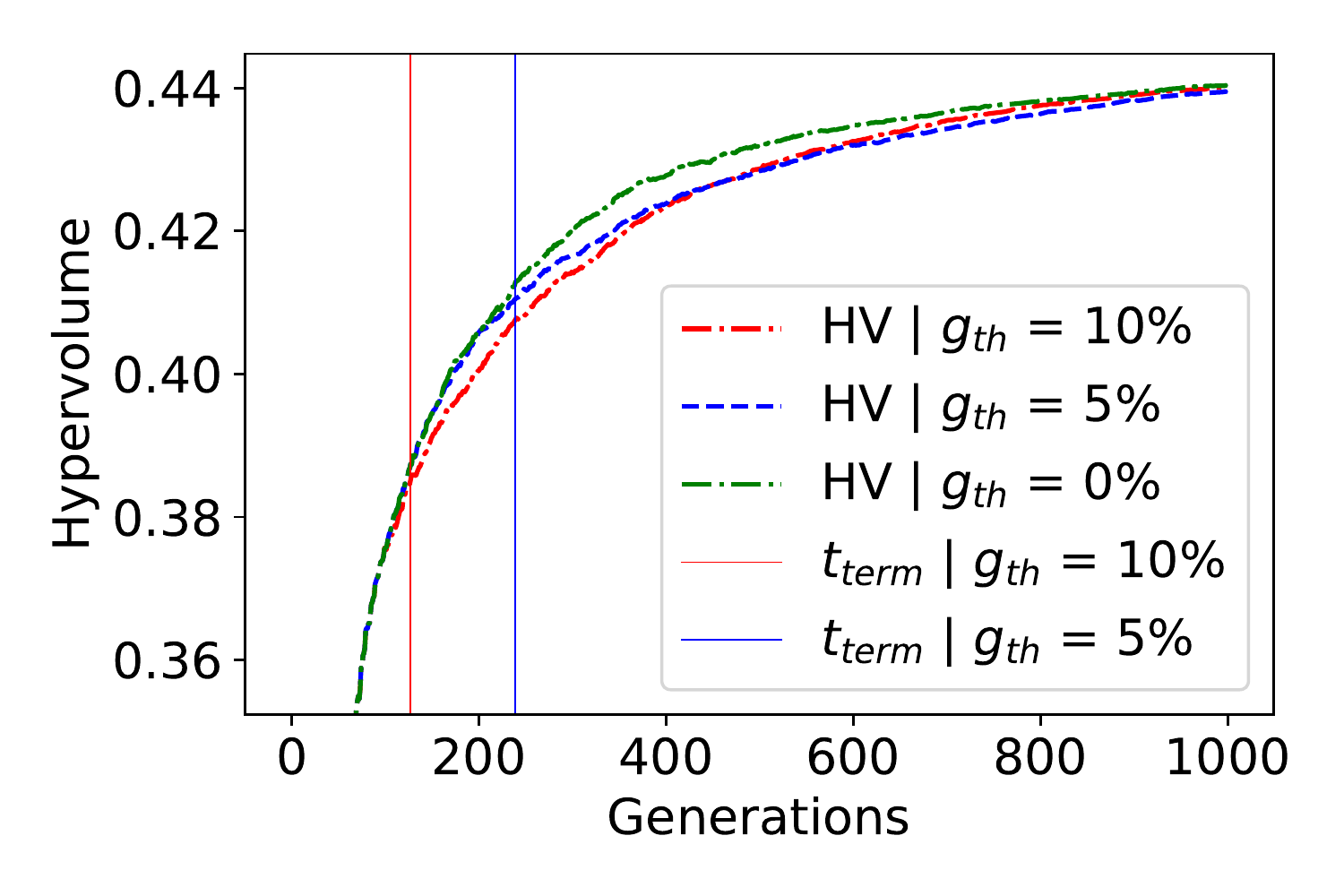}
    \caption{WFG7 ($M=3$)}
    \label{fig:HV-WFG7-term}
\end{subfigure}
\caption{Limited learning on two WFG problems.}
\label{fig:WFG-termination}
\end{figure}

In this subsection, the effect of terminating the RF-based learning process with a threshold $g_{th}$ and its importance are discussed. Figure~\ref{fig:WFG-termination} shows the results of varying $g_{th}$ to be 10\%, 5\% and 0\% (no termination) on some problems. It can be seen that all three variants are the same until the vertical red line, where learning is terminated and $g_{th}=10\%$ starts showing a little worse performance than the others. Similarly, after the vertical blue line, the one with $g_{th}=5\%$ starts performing worse. However, it may be noted that the performance difference is only minor. This is because the learning-based repair does not provide large improvements as the solution nears the Pareto-optimal set. Additional repair will reduce diversity in non-dominated solutions and may not help the search process. Thus, terminating the learning-based repair operator allows original EMO/EMaO algorithms to work well and additionally saves the extra computational effort spent on learning the IR2 operator. Similar trends are observed in other problems as well. However, the early terminations depicted here clearly occur before the beneficial effects of the learning and repair operation are exhausted. At 300-800 generations on either function, the results of continuing learning/repair are found to be beneficial.

\begin{figure}[hbt]    
\begin{subfigure}[b]{0.49\linewidth}
    \centering    
    \includegraphics[width=1.05\linewidth]{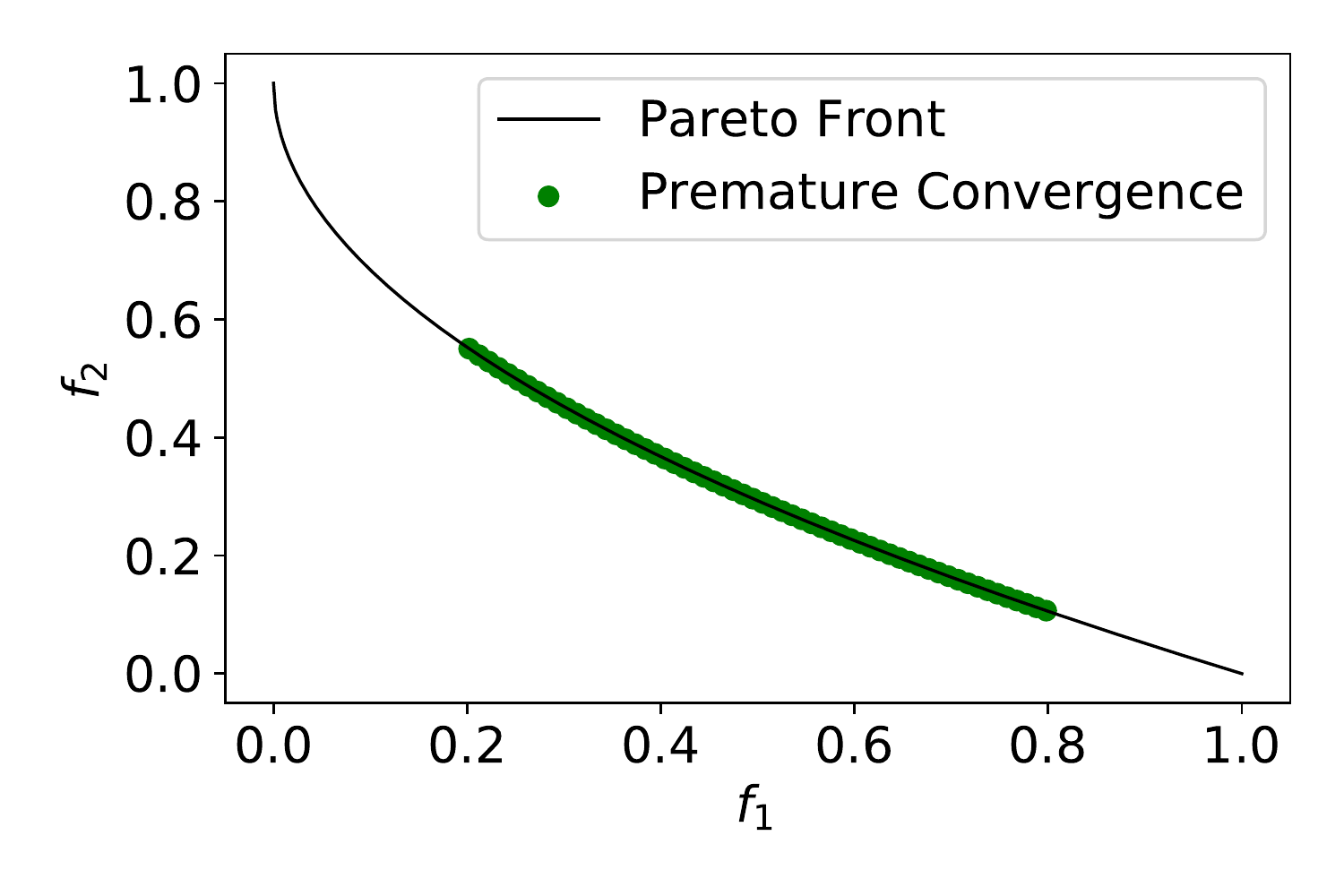}
    \caption{Premature and final Front}
    \label{fig:Front-CIBN2}
    \end{subfigure}
\begin{subfigure}[b]{0.49\linewidth}
    \centering
    \includegraphics[width=1.05\linewidth]{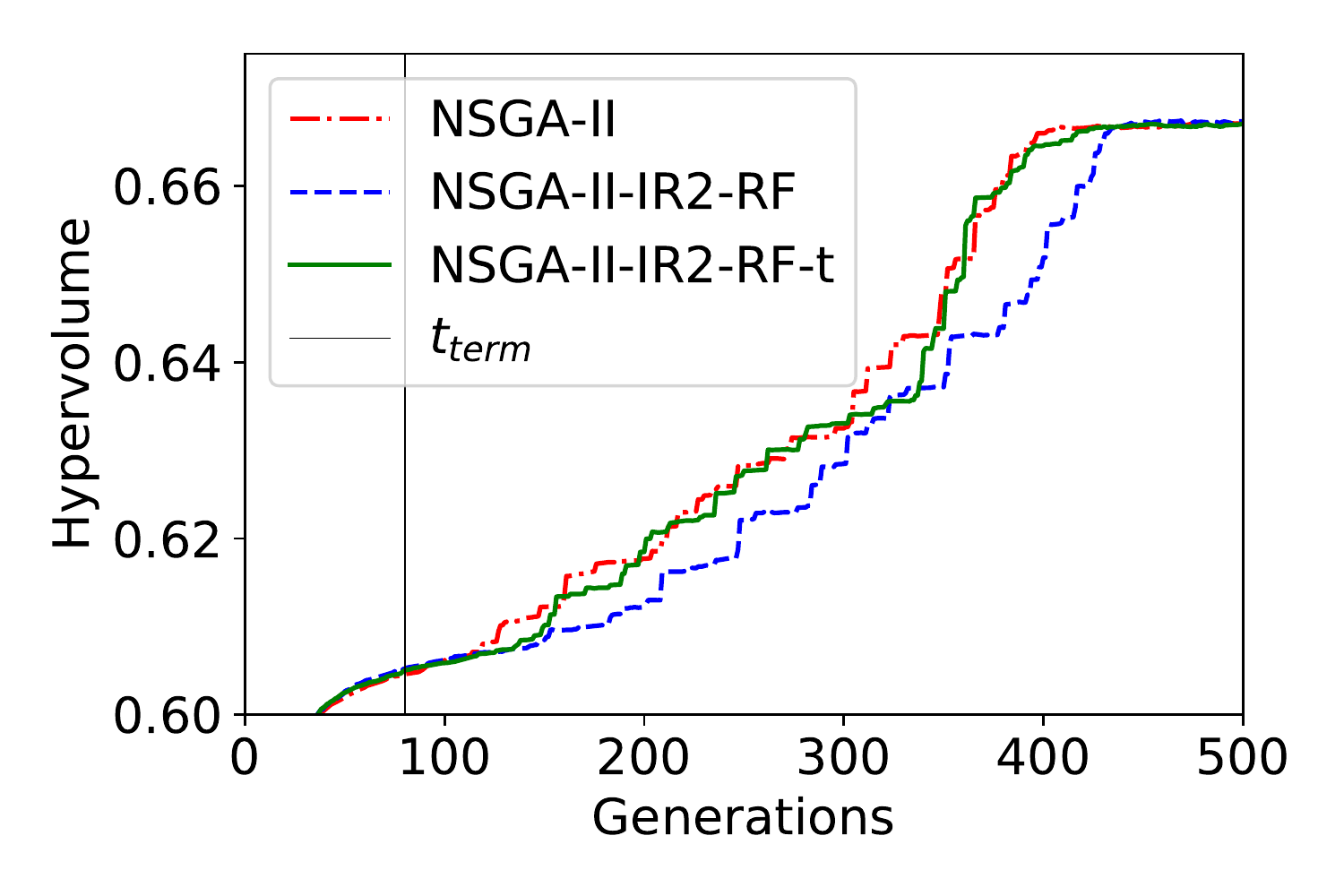}
    \caption{Hypervolume plot}
    \label{fig:HV-CIBN2}
\end{subfigure}
\caption{Analysis on constrained imbalance problem CIBN2.}
\label{fig:CIBN2}
\end{figure}

There can be some problems where the idea of termination becomes vital, where the algorithm first converges to only a portion of the PO front and then spreads along it.  For example, the constrained imbalance problem CIBN2 \cite{CIBN-2017}, for which the PO front is shown with a solid black line in Figure~\ref{fig:Front-CIBN2} is such a problem. The green dots represent the premature front where the optimizer converges first. The proposed IR2 operator may not help in this case (after the green-part in Figure~\ref{fig:Front-CIBN2} is achieved), since there are no representative solutions in other parts of the front to learn from. Also, the repair operator (which learns from the existing good solutions only) may also repair some potentially good offspring in the unexplored regions back to the existing front, which is not favorable. Hence, terminating the repair operator can help in reducing the chances of such performance degradation. Figure~\ref{fig:HV-CIBN2} shows the median hypervolume plot as the solutions evolve from the premature front to the PO front. It can be seen that the NSGA-II-IR2-RF (without termination) performs worse than NSGA-II, but that NSGA-II-IR2-RF with termination (NSGA-II-IR2-t) performs at par with NSGA-II.

\section{Conclusions} \label{sec:conclusions}

In this paper, we have proposed an enhanced \textit{Innovized} repair operator, named IR2, for potentially improving some of the current offspring solutions, so they become equivalent to or better than the current non-dominated solutions. To fulfill this aim, we have first paired the (archived) solutions from the past few generations with the \textit{best} non-dominated solution found till the current generation, along each reference vector. The dataset constituted by such paired solutions is  used to train a Random Forest model which in turn is utilized, as a repair operator. The latter ensures transition of an offspring $X$ to a better converged solution $Y$.

The proposed IR2 operator has been integrated with NSGA-II, NSGA-III and MOEA/D and tested on multiple benchmark problems ranging from two to five objectives, a total of 63 test instances. The repair shows significant improvements in earlier generations, while the performance is either better or at par towards the end, as compared with the base EMO/EMaO algorithm. Later, this paper has discussed certain characteristics of problems where the proposed IR2 operator yields good results, such as multi-modality, bias, and inter-variable linkages. A restricted use of the learning approach has been found to reduce the computational expense and to improve convergence behavior in certain problems. 

This paper has also discussed the effect of boundary points (and interior points) in dealing with many-objective optimization problems. With a recently proposed energy method \cite{Energy-TEC-2020}, we have shown that decreasing the proportion of boundary points or increasing the proportion of interior points while keeping the population size similar can help improve the learning process, and hence, accelerate convergence. In future studies, this work can be extended to single-objective multi-modal problems and to discrete- and mixed-variable problems. The effect of distribution of reference points on quality of learning can be explored. It would also be interesting to divide the variables into different ``buckets'' based on their dependencies, since the machine learning algorithms would then work better with fewer variables. Nevertheless, this extensive study has revealed a new concept using an AI/ML method to improve the performance of EMO/EMaO algorithms for solving complex optimization problems. 



%



\section*{Acknowledgment}

Authors would like to acknowledge the support provided by the Government of India under SPARC project No. P66. The authors would also thank Blank and Deb \cite{pymoo-paper} for developing the python-based pymoo platform which has been utilized in this research.

\ifCLASSOPTIONcaptionsoff
  \newpage
\fi



%

\bibliographystyle{IEEETran}
\bibliography{IEEEabrv,car-ref}

\begin{thebibliography}{10}
\providecommand{\url}[1]{#1}
\csname url@samestyle\endcsname
\providecommand{\newblock}{\relax}
\providecommand{\bibinfo}[2]{#2}
\providecommand{\BIBentrySTDinterwordspacing}{\spaceskip=0pt\relax}
\providecommand{\BIBentryALTinterwordstretchfactor}{4}
\providecommand{\BIBentryALTinterwordspacing}{\spaceskip=\fontdimen2\font plus
\BIBentryALTinterwordstretchfactor\fontdimen3\font minus
  \fontdimen4\font\relax}
\providecommand{\BIBforeignlanguage}[2]{{%
\expandafter\ifx\csname l@#1\endcsname\relax
\typeout{** WARNING: IEEEtran.bst: No hyphenation pattern has been}%
\typeout{** loaded for the language `#1'. Using the pattern for}%
\typeout{** the default language instead.}%
\else
\language=\csname l@#1\endcsname
\fi
#2}}
\providecommand{\BIBdecl}{\relax}
\BIBdecl

\bibitem{deb-book-01}
K.~Deb, \emph{Multi-objective optimization using evolutionary
  algorithms}.\hskip 1em plus 0.5em minus 0.4em\relax Chichester, UK: Wiley,
  2001.

\bibitem{carlos-book}
C.~A.~C. Coello, D.~A. VanVeldhuizen, and G.~Lamont, \emph{Evolutionary
  Algorithms for Solving Multi-Objective Problems}.\hskip 1em plus 0.5em minus
  0.4em\relax Boston, MA: Kluwer, 2002.

\bibitem{deb2002nsga1}
K.~Deb, A.~Pratap, S.~Agarwal, and T.~Meyarivan, ``A fast elitist non-dominated
  sorting genetic algorithm for multi-objective optimization: Nsga-ii,''
  \emph{IEEE transactions on evolutionary computation}, vol.~6, no.~2, pp.
  182--197, 2002.

\bibitem{nsga3-1}
K.~Deb and H.~Jain, ``An evolutionary many-objective optimization algorithm
  using reference-point based non-dominated sorting approach, {Part I: Solving}
  problems with box constraints,'' \emph{IEEE Transactions on Evolutionary
  Computation}, vol.~18, no.~4, pp. 577--601, 2014.

\bibitem{nsga3-2}
H.~Jain and K.~Deb, ``An evolutionary many-objective optimization algorithm
  using reference-point based non-dominated sorting approach, {Part II:
  Handling} constraints and extending to an adaptive approach,'' \emph{IEEE
  Transactions on Evolutionary Computation}, vol.~18, no.~4, pp. 602--622,
  2014.

\bibitem{moead}
Q.~Zhang and H.~Li, ``{MOEA/D: A} multiobjective evolutionary algorithm based
  on decomposition,'' \emph{Evolutionary Computation, IEEE Transactions on},
  vol.~11, no.~6, pp. 712--731, 2007.

\bibitem{boa}
M.~Pelikan, D.~E. Goldberg, and E.~Cantú-Paz, ``The bayesian optimization
  algorithm,'' in \emph{GECCO}.\hskip 1em plus 0.5em minus 0.4em\relax Morgan
  Kaufmann Publishers Inc., 1999, pp. 525--532.

\bibitem{MONEDA}
M.~Luis, G.~Jesus, B.~Antonio, and M.~J. Manuel, ``Introducing moneda: scalable
  multiobjective optimization with a neural estimation of distribution
  algorithm,'' in \emph{Proceedings of Genetic and Evolutionary Computation
  conference (GECCO-2008)}, 2008, pp. 689--696.

\bibitem{sukrit-coin-ann}
\BIBentryALTinterwordspacing
S.~Mittal, D.~Saxena, K.~Deb, and E.~Goodman, ``An {ANN}--assisted repair
  operator for evolutionary multiobjective optimization,'' Department of
  Electrical and Computer Engineering, Michigan State University, East Lansing,
  USA, Tech. Rep. COIN Report No. 2020005, 2020. [Online]. Available:
  \url{https://www.egr.msu.edu/~kdeb/papers/c2020005.pdf}
\BIBentrySTDinterwordspacing

\bibitem{abhinav-cec17}
A.~Gaur and K.~Deb, ``Effect of size and order of variables in rules for
  multi-objective repair-based innovization procedure,'' in \emph{Proceedings
  of Congress on Evolutionary Computation (CEC-2017) Conference}.\hskip 1em
  plus 0.5em minus 0.4em\relax Piscatway, NJ: IEEE Press, 2017.

\bibitem{deb-datta-ejor}
K.~Deb and R.~Datta, ``Hybrid evolutionary multi-objective optimization and
  analysis of machining operations,'' \emph{Engineering Optimization}, vol.~44,
  no.~6, pp. 685--706, 2012.

\bibitem{rocket-cec20}
A.~{Ghosh}, E.~{Goodman}, K.~{Deb}, R.~{Averill}, and A.~{Diaz}, ``A
  large-scale bi-objective optimization of solid rocket motors using
  innovization,'' in \emph{2020 IEEE Congress on Evolutionary Computation
  (CEC)}, 2020, pp. 1--8.

\bibitem{eopaper}
S.~Bandaru and K.~Deb, ``Towards automating the discovery of certain innovative
  design principles through a clustering based optimization technique,''
  \emph{Engineering optimization}, vol.~43, no.~9, pp. 911--941, 2011.

\bibitem{dudas2013integration}
C.~Dudas, A.~H. Ng, L.~Pehrsson, and H.~Bostr{\"o}m, ``Integration of data
  mining and multi-objective optimisation for decision support in production
  systems development,'' \emph{International Journal of Computer Integrated
  Manufacturing}, no. ahead-of-print, pp. 1--16, 2013.

\bibitem{bandaru-gp}
S.~Bandaru and K.~Deb, ``A dimensionally-aware genetic programming architecture
  for automated innovization,'' in \emph{Proceedings of the Seventh
  International Conference on Evolutionary Multi-Criterion Optimization
  (EMO-13), LNCS 7811}.\hskip 1em plus 0.5em minus 0.4em\relax Heidelberg:
  Springer, 2013, pp. 513--527.

\bibitem{sukrit-ann-repair-gecco}
\BIBentryALTinterwordspacing
S.~Mittal, D.~K. Saxena, and K.~Deb, ``Learning-based multi-objective
  optimization through {ANN}--assisted online innovization,'' in
  \emph{Proceedings of the 2020 Genetic and Evolutionary Computation Conference
  Companion}, ser. GECCO 20.\hskip 1em plus 0.5em minus 0.4em\relax New York,
  NY, USA: Association for Computing Machinery, 2020, pp. 171--172. [Online].
  Available: \url{https://doi.org/10.1145/3377929.3389925}
\BIBentrySTDinterwordspacing

\bibitem{survey_prob}
M.~Pelikan, D.~Goldberg, and F.~Lobo, ``A survey of optimization by building
  and using probabilistic models,'' \emph{Computational Optimization and
  Applications}, vol.~21, no.~1, pp. 5--20, 2002.

\bibitem{eda_svm}
L.~Li, H.~Chen, and C.~L. et~al., ``A robust hybrid approach based on
  estimation of distribution algorithm and support vector machine for hunting
  candidate disease genes,'' \emph{The Scientific World Journal}, vol. 2013,
  no. 393570, p.~7, 2013.

\bibitem{repair-survey}
S.~Salcedo-Sanz, ``A survey of repair methods used as constraint handling
  techniques in evolutionary algorithms,'' \emph{Computer Science Review},
  vol.~3, no.~3, pp. 175 -- 192, 2009.

\bibitem{memetic-de}
C.~Zhang, J.~Chen, and B.~Xin, ``Distributed memetic differential evolution
  with the synergy of lamarckian and baldwinian learning,'' \emph{Applied Soft
  Computing}, vol.~13, no.~5, pp. 2947 -- 2959, 2013.

\bibitem{MIAB}
Y.~Qi, F.~Liu, M.~Liu, M.~Gong, and L.~Jiao, ``Multi-objective immune algorithm
  with baldwinian learning,'' \emph{Applied Soft Computing}, vol.~12, no.~8,
  pp. 2654 -- 2674, 2012.

\bibitem{deb-design}
K.~Deb, ``Unveiling innovative design principles by means of multiple
  conflicting objectives,'' \emph{Engineering Optimization}, vol.~35, no.~5,
  pp. 445--470, 2003.

\bibitem{deb-innovization}
K.~Deb and A.~Srinivasan, ``Innovization: Innovating design principles through
  optimization.'' in \emph{Proceedings of the Genetic and Evolutionary
  Computation Conference (GECCO-2006)}, New York: ACM, 2006, pp. 1629--1636.

\bibitem{das}
I.~Das and J.~Dennis, ``Normal-boundary intersection: {A} new method for
  generating the {P}areto surface in nonlinear multicriteria optimization
  problems,'' \emph{SIAM Journal of Optimization}, vol.~8, no.~3, pp. 631--657,
  1998.

\bibitem{unit-simplex-emo}
K.~Deb, S.~Bandaru, and H.~Seada, ``Generating uniformly distributed points on
  a unit simplex for evolutionary many-objective optimization,'' in
  \emph{Proceedings of the Tenth International Conference on Evolutionary
  Multi-Criterion Optimization (EMO-19), LNCS 11411}.\hskip 1em plus 0.5em
  minus 0.4em\relax Heidelberg: Springer, 2019, pp. 179--190.

\bibitem{wierzbicki}
A.~P. Wierzbicki, ``The use of reference objectives in multiobjective
  optimization,'' in \emph{Multiple Criteria Decision Making Theory and
  Applications}, G.~Fandel and T.~Gal, Eds.\hskip 1em plus 0.5em minus
  0.4em\relax Berlin: Springer-Verlag, 1980, pp. 468--486.

\bibitem{boundary-PSO}
N.~Padhye, K.~Deb, and P.~Mittal, ``Boundary handling approaches in particle
  swarm optimization,'' in \emph{Proceedings of Seventh International
  Conference on Bio-Inspired Computing: Theories and Applications (BIC-TA
  2012). Advances in Intelligent Systems and Computing}, J.~Bansal, P.~Singh,
  K.~Deep, M.~Pant, and A.~Nagar, Eds., vol. 201.\hskip 1em plus 0.5em minus
  0.4em\relax Springer, India, 2013, pp. 287--298.

\bibitem{ZDT}
\BIBentryALTinterwordspacing
E.~Zitzler, K.~Deb, and L.~Thiele, ``Comparison of multiobjective evolutionary
  algorithms: Empirical results,'' \emph{Evolutionary Computation}, vol.~8,
  no.~2, pp. 173--195, 2000. [Online]. Available:
  \url{https://doi.org/10.1162/106365600568202}
\BIBentrySTDinterwordspacing

\bibitem{WFG}
S.~{Huband}, P.~{Hingston}, L.~{Barone}, and L.~{While}, ``A review of
  multiobjective test problems and a scalable test problem toolkit,''
  \emph{IEEE Transactions on Evolutionary Computation}, vol.~10, no.~5, pp.
  477--506, 2006.

\bibitem{DTLZ}
\BIBentryALTinterwordspacing
K.~Deb, L.~Thiele, M.~Laumanns, and E.~Zitzler, \emph{Scalable Test Problems
  for Evolutionary Multiobjective Optimization}.\hskip 1em plus 0.5em minus
  0.4em\relax London: Springer London, 2005, pp. 105--145. [Online]. Available:
  \url{https://doi.org/10.1007/1-84628-137-7\_6}
\BIBentrySTDinterwordspacing

\bibitem{NSGA-II-III-comparison}
H.~{Ishibuchi}, R.~{Imada}, Y.~{Setoguchi}, and Y.~{Nojima}, ``Performance
  comparison of nsga-ii and nsga-iii on various many-objective test problems,''
  in \emph{2016 IEEE Congress on Evolutionary Computation (CEC)}, July 2016,
  pp. 3045--3052.

\bibitem{Energy-TEC-2020}
J.~Blank, K.~Deb, Y.~Dhebar, S.~Bandaru, and H.~Seada, ``Generating well-spaced
  points on a unit simplex for evolutionary many-objective optimization,''
  \emph{IEEE Transactions on Evolutionary Computation}, pp. 1--1, 2020.

\bibitem{CIBN-2017}
J.~{Lin}, H.~{Liu}, and C.~{Peng}, ``The effect of feasible region on
  imbalanced problem in constrained multi-objective optimization,'' in
  \emph{2017 13th International Conference on Computational Intelligence and
  Security (CIS)}, 2017, pp. 82--86.

\bibitem{pymoo-paper}
J.~{Blank} and K.~{Deb}, ``Pymoo: Multi-objective optimization in python,''
  \emph{IEEE Access}, vol.~8, pp. 89\,497--89\,509, 2020.

\end{thebibliography}



%




\end{document}